\documentclass[10pt,twocolumn,letterpaper]{article}

\usepackage{cvpr}              

\usepackage[accsupp]{axessibility}  
\definecolor{cvprblue}{rgb}{0.21,0.49,0.74}
\usepackage[pagebackref,breaklinks,colorlinks,allcolors=cvprblue]{hyperref}
\usepackage[table]{xcolor}  

\def\confName{CVPR}
\def\confYear{2026}

\title{DreamStereo: Towards Real-Time Stereo Inpainting for HD Videos}
\author{
\textnormal{
Yuan Huang$^{*}$ \quad
Sijie Zhao$^{*}$ \quad
Jing Cheng \quad
Hao Xu \quad
Shaohui Jiao$^{\dagger}$
}\\
ByteDance\\
{\tt\small \{huangyuan.21, zhaosijie, chengjing.amber, xuhao.sky, jiaoshaohui\}@bytedance.com}
}

\begin{document}
\maketitle
\begingroup
\renewcommand{\thefootnote}{\fnsymbol{footnote}}
\setcounter{footnote}{0}
\footnotetext[1]{Equal contribution.}
\footnotetext[2]{Corresponding author.}
\footnotetext[3]{Project page: \url{https://huangyuan2020.github.io/DreamStereo/}.}
\endgroup

\begin{abstract}
Stereo video inpainting, which aims to fill the occluded regions of warped videos with visually coherent content while maintaining temporal consistency, remains a challenging open problem. 
The regions to be filled are scattered along object boundaries and occupy only a small fraction of each frame, leading to two key challenges. 
First, existing approaches perform poorly on such tasks due to the scarcity of high-quality stereo inpainting datasets, which limits their ability to learn effective inpainting priors. 
Second, these methods apply equal processing to all regions of the frame, even though most pixels require no modification, resulting in substantial redundant computation. 
To address these issues, we introduce three interconnected components. 
We first propose Gradient-Aware Parallax Warping (GAPW), which leverages backward warping and the gradient of the coordinate mapping function to obtain continuous edges and smooth occlusion regions. 
Then, a Parallax-Based Dual Projection (PBDP) strategy is introduced, which incorporates GAPW to produce geometrically consistent stereo inpainting pairs and accurate occlusion masks without requiring stereo videos. 
Finally, we present Sparsity-Aware Stereo Inpainting (SASI), which reduces over 70\% of redundant tokens, achieving a 10.7$\times$ speedup during diffusion inference and delivering results comparable to its full-computation counterpart, enabling real-time processing of HD ($768\times1280$) videos at 25\,FPS on a single A100 GPU.

\end{abstract}
\vspace{-8pt}

\section{Introduction}
\label{sec:intro}

With the advancement of AR/VR devices, stereo digital media are experiencing rapid growth and gaining increasing popularity. These media deliver a more immersive and enjoyable experience by projecting visual content with parallax to different eyes. However, the high cost of multi-camera visual capture equipment hinders the broader development of stereo visual content.  Currently, monocular visual content far exceeds its stereo counterpart, as monocular display devices such as smartphones, tablets, and computers remain the mainstream.  Thus, a question emerges: How to convert the vast amount of monocular visual content available into stereo counterparts?

Human beings perceive the depth of the 3D world through the parallax between the left and right eyes. Early 2D-to-3D methods, such as Depth Image Based Rendering (DIBR)~\cite{zhang20113d, konrad2013learning}, first predict the depth of the input view and then render it to another view using camera intrinsic and extrinsic parameters. Unfortunately, these methods perform poorly because they fail to fill the occluded areas that often appear at the edges of objects in the other view. Multiplane Images (MPI) methods~\cite{tucker2020single, han2022single, zhang2023structural} decompose visual content into different layers along the depth dimension and perform inpainting on each background layer. However, they struggle when objects within the same layer have large disparity.

\begin{figure}[t]
\includegraphics[width=\columnwidth]{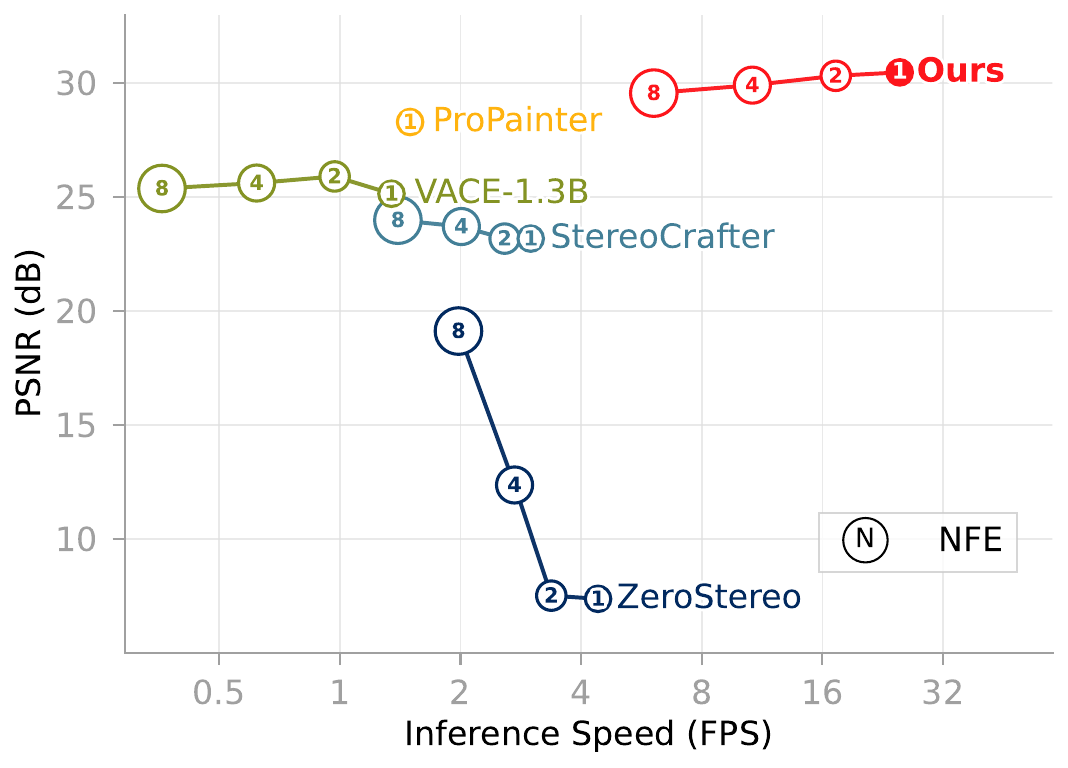}
\caption{Comparison among different methods at a resolution of $768\times 1280$: our method achieves a PSNR of 30.5 dB at 1 NFE, with an inference speed of 24.9 fps.}
\label{fig:cover}
\vspace{-5pt}
\end{figure}

Methods~\cite{zhao2024stereocrafter, yu2025trajectorycrafter, zhang2024spatialme} that project images or videos based on the depth and then use diffusion models to inpaint occluded regions have gained growing popularity. A key aspect of such methods is the construction of high-quality stereo inpainting datasets. StereoCrafter~\cite{zhao2024stereocrafter}, ImmersePro~\cite{shi2024immersepro}, StereoConversion~\cite{mehl2024stereo} collect stereo images/videos and use stereo matching~\cite{jing2024match, wen2025foundationstereo} to establish the disparity between the views of the left and right eyes. However, acquiring stereo videos is typically expensive, and different stereo videos may adopt varying binocular projection rules, or even simple pixel translation, which degrades data quality. TrajectoryCrafter~\cite{yu2025trajectorycrafter} proposes building novel view datasets from monocular videos through double reprojection, but its use of forward warping leads to scattered pixels at the edges of objects, compromising dataset quality and significantly degrading inpainting performance during inference due to incorrect foreground pixels. 

Another issue is that occlusions only account for a small portion of each frame. However, all the aforementioned stereo inpainting methods~\cite{zhao2024stereocrafter, yu2025trajectorycrafter, zhang2024spatialme, shi2024immersepro, mehl2024stereo, wang2025zerostereo} process all pixels uniformly, although most pixels do not require modification, leading to an extremely slow inference speed. This prompts us to question whether it is necessary for all visual tokens to participate in computation. In Diffusion Transformers (DiT)~\cite{peebles2023scalable}, the computational complexity of attention can be expressed as $\mathcal{O}(N^2)$, where $N$ denotes the number of visual tokens. This implies that when the number of tokens decreases, the inference speed of DiT will be rapidly improved, which is highly appealing.

To address the issue of scattered pixels and masks caused by forward warping, we propose Gradient-Aware Parallax Warping, a method that combines backward warping and the gradients of coordinates mapping function to obtain smooth and accurate pixels and occlusion masks for the warped video. Based on this proposed warping method, we introduce Parallax-Based Dual Projection, a strategy that generates stereo inpainting datasets utilizing massive monocular video data, which can be easily scaled up compared to the data construction methods based on stereo videos. Finally and most importantly, we present Sparsity-Aware Stereo Inpainting and review the redundancy of visual tokens in such task. By reducing more than 70\% of visual tokens, we improve the inference speed of DiT by $10.7\times$, pushing the processing of HD videos in stereo inpainting into the real-time era. Our main contributions are summarized as follows,

\begin{itemize}

\item We propose Gradient-Aware Parallax Warping (GAPW), a method to warp images or videos to novel views with smoother masks and fewer scattered pixels.

\item We introduce Parallax-Based Dual Projection (PBDP), a strategy for generating stereo inpainting data without the need for stereo videos.

\item We present Sparsity-Aware Stereo Inpainting (SASI), a real-time stereo inpainting method for HD videos by eliminating the redundancy of visual tokens.

\end{itemize}

\vspace{12pt}

\section{Related Work}

\paragraph{Novel View Synthesis}
Novel View Synthesis has been a widely studied topic in 3D vision. Early DIBR-based methods~\cite{zhang20113d, konrad2013learning} render undistorted novel views by utilizing depth and camera pose, yet they are ineffective in handling occluded regions in new perspectives. Later, MPI-Based methods~\cite{tucker2020single, zhang2023structural, han2022single} were proposed to decompose visual content into different layers along the depth dimension and render novel views through inpainting background layers. However, they fail to deal with objects with large disparities in a single layer. With the rise of 3D representations, NeRF~\cite{mildenhall2021nerf, hong2022headnerf, pumarola2021d} and 3DGS~\cite{kerbl20233d, wu20244d} have been applied to novel view synthesis. These methods usually require frame-by-frame calibration of camera poses for videos, performing poorly when there are large camera and object movements. Moreover, they often need separate optimization for each scene, which reduces efficiency. Methods like CAT3D~\cite{gao2024cat3d} and ViewCrafter~\cite{yu2024viewcrafter} achieve the generation of novel views of static objects from single or multiple images by combining diffusion models~\cite{rombach2021highresolution, blattmann2023stable} with NeRF. SV4D~\cite{xie2024sv4d} and CAT4D~\cite{xie2024sv4d} further integrate video diffusion~\cite{blattmann2023stable, yang2024cogvideox} to generate 4D scenes from single-video inputs. Nevertheless, these methods generally focus on large camera baselines and tend to produce blurry backgrounds, making them difficult to apply to stereo video generation. TrajectoryCrafter~\cite{yu2025trajectorycrafter} proposes a double reprojection strategy to produce novel view inpainting data through monocular videos, but the used forward warping causes scattered pixels at object edges, affecting dataset quality and reducing inpainting performance.
\vspace{-5pt}

\paragraph{Mono-to-Stereo Conversion}
In the field of mono-to-stereo conversion, early work such as Deep3D~\cite{xie2016deep3d} trained on stereo pairs to directly predict the right view from the left input, but the generated results often exhibit notable disparity artifacts. 
Recent advances mainly rely on diffusion models, leveraging their strong priors to produce visually consistent results. 
Training-free methods such as StereoDiffusion~\cite{wang2024stereodiffusion}, SVG~\cite{dai2024svg}, and ZeroStereo~\cite{wang2025zerostereo} generate stereo images or videos by modifying latent variables during denoising, though this may cause domain gaps during inference. 
Training-based approaches, including StereoCrafter~\cite{zhao2024stereocrafter}, Restereo~\cite{huang2025restereo}, ImmersePro~\cite{shi2024immersepro}, SpatialMe~\cite{zhang2024spatialme}, and StereoConversion~\cite{mehl2024stereo}, typically formulate 2D-to-3D generation as a stereo inpainting task via forward warping. 
More recent diffusion-based efforts such as SpatialDreamer~\cite{2024spatialdreamer} and M2SVid~\cite{shvetsova2025m2svid} further explore stereo and 4D generation tasks. 
However, most existing models rely on stereo datasets, whose collection is costly and inconsistent due to varying binocular projection settings. 
Moreover, none of the diffusion-based methods considers the intrinsic sparsity of stereo inpainting, leaving real-time performance unattainable.

\section{Method}
 In this section, we first introduce Gradient-Aware Parallax Warping (GAPW), a novel view synthesis technique that reduces fly-point artifacts and yields cleaner occlusion boundaries compared to the commonly used forward warping. Subsequently, we present Parallax-Based Dual Projection (PBDP), a strategy for generating high-quality stereo inpainting data directly from monocular videos, without requiring ground-truth stereo supervision. Finally, we propose Sparsity-Aware Stereo Inpainting (SASI), a real-time stereo inpainting method for HD videos by eliminating the redundancy of visual tokens.

 \subsection{Gradient-Aware Parallax Warping}
 \label{sec:gapw}
 In novel view synthesis, the goal is to transform a source image $ I(x, y)$ with coordinates $  (x, y) \in \Omega \subseteq \mathbb{Z}^2  $ into a new view $ I'(x', y') $ where $ (x', y') \in \Omega' \subseteq \mathbb{Z}^2  $. Most existing methods~\cite{zhao2024stereocrafter, yu2025trajectorycrafter, huang2025restereo} achieve this conversion by forward warping~\cite{Niklaus_CVPR_2020}. Let the transform function $  T: \mathbb{R}^2 \to \mathbb{R}^2  $ denote the mapping from $(x, y)$ to $(x', y')$, 
\vspace{-6pt}

\begin{equation}
(x', y') = T(x, y; D, C)
\label{eq1}
\end{equation}

Here, $  D  $ represents the disparity of the source image, and $  C  $ denotes the camera pose. Once the camera pose is fixed and the disparity is estimated, the function $T$ is uniquely determined. Subsequently, $  I(x, y)  $ is assigned to the corresponding position $  (x', y')  $ in the target image as,
\vspace{-6pt}

\begin{equation}
I'(x', y') \leftarrow I(x, y)
\end{equation}

\begin{figure}[t]
\centering
\includegraphics[width=\columnwidth]{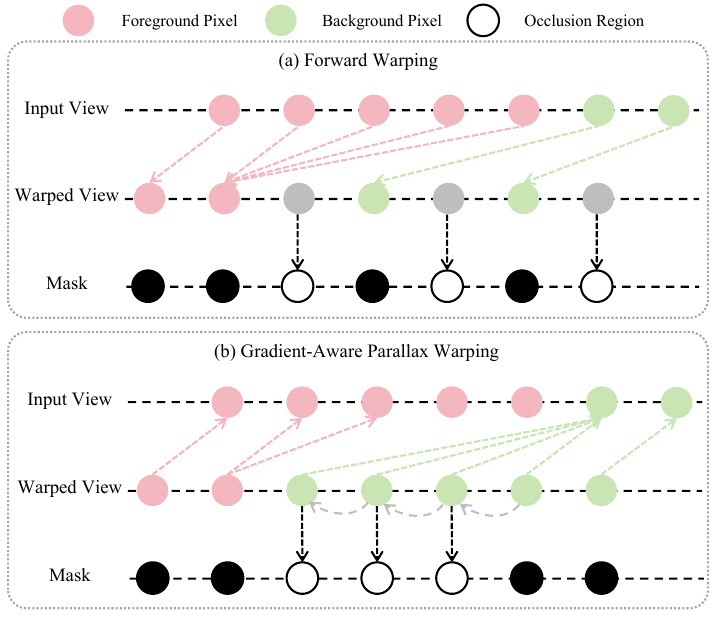} 
\caption{Comparison of forward warping and our GAPW in pixel mapping and occlusion mask generation.}
\label{fig:parallax_warping}
\vspace{-12pt}
\end{figure}

This method generates holes in regions not assigned by any source pixels, i.e., occluded areas, which facilitates subsequent inpainting. 
However, it often produces misaligned artifacts in multi-layer background regions, possibly related to depth-scale inconsistencies across layers, thereby degrading both stereo dataset construction and inpainting performance.
The simplified illustration of this process is shown in Fig.~\ref{fig:parallax_warping}(a).

To address this issue, we incorporate backward warping~\cite{mcguire2005steep} with gradient in consideration. For each $T$, there exists an inverse function $  T^{-1}: \mathbb{R}^2 \to \mathbb{R}^2  $, such that,

\begin{equation}
(x, y) = T^{-1}(x', y'; D, C)
\end{equation}

The pixel value of the target image at $  (x', y')  $ can be obtained through interpolation:

\begin{equation}
I'(x', y') = \text{Interpolate}(I, x, y)
\end{equation}

We then calculate the gradients of the $ (x', y') $ with respect to the $  (x, y)  $ based on Eq.~\ref{eq1},

\begin{equation}
\mathbf{J}_T(x', y')= 
\begin{pmatrix}
    \mathbf{\nabla} x' \\ \mathbf{\nabla} y'
\end{pmatrix} = \begin{pmatrix} 
    \frac{\partial x'}{\partial x} & \frac{\partial x'}{\partial y} \\ 
    \frac{\partial y'}{\partial x} & \frac{\partial y'}{\partial y} 
\end{pmatrix}
\end{equation}

The gradients can be represented by the Jacobian Matrix $  \mathbf{J}_T(x', y')  $, which describes the deformation of pixels at position $  (x', y')  $ in x- and y- directions. The occluded pixels in the new view occur in regions with larger deformation. Thus, the occlusion mask $  M(x', y')$ can be obtained by a threshold $\delta$,

\begin{equation}
    M(x', y') ~=~ \parallel \mathbf{J}_T(x', y') \parallel_2 ~>~ \delta
\end{equation}

In the binocular system, pixels undergo warping only in the x-direction, thus can be simplified as,

\begin{equation}
    M(x', y') ~=~ \left|  \frac{\partial x'}{\partial x} \right| ~>~ \delta
\end{equation}

Finally, we illustrate the calculation of the warped view and mask under the binocular system in Fig.~\ref{fig:parallax_warping}(b). Due to the continuity of the gradient, our method is capable of generating smooth occluded regions and avoiding the artifacts of scattered pixels.

\begin{figure*}[t]
\centering
\includegraphics[width=\textwidth]{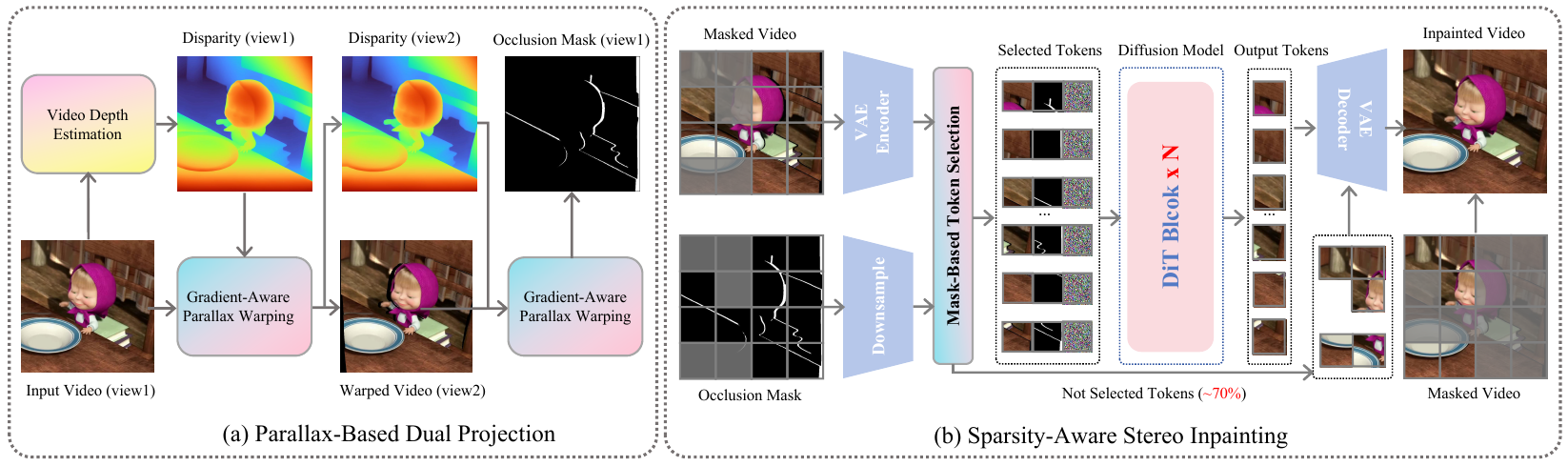} 
\caption{(\textbf{a}) Illustration of the Parallax-Based Dual Projection, which utilizes Gradient-Aware Parallax Warping for reprojection to obtain the occlusion mask under input view. (\textbf{b}) Our proposed Sparsity-Aware Stereo Inpainting utilizes Mask-Based Token Selection to reduce the redundancy of visual tokens.}
\label{fig：framework}
\vspace{-12pt}
\end{figure*}

\subsection{Parallax-Based Dual Projection}
\label{sec:pbdp}
Inspired by TrajectoryCrafter~\cite{yu2025trajectorycrafter}, we use a reprojection strategy to produce stereo inpainting pairs without the help of stereo videos. As illustrated in Fig.~\ref{fig：framework}(a), given a monocular video $\mathbf{V}_1$ in \texttt{view1}, we first use a depth estimator (e.g., DepthCrafter~\cite{hu2025-DepthCrafter}) to predict the disparity $\mathbf{D}_1$.  Then we synthesize a warped video $V_2$ and its disparity map $\mathbf{D}_2$ in \texttt{view2} with the help of GAPW,

\begin{equation}
\mathbf{V}_2, \mathbf{D}_2 = \text{GAPW}(\mathbf{V}_1, \mathbf{D}_1; v1\rightarrow v2),
\end{equation}
where $v1\rightarrow v2$ means warping from \texttt{view1} to \texttt{view2}. Subsequently, we apply the GAPW for $V_2$ to reconstruct the $V_1'$ and obtain the occlusion mask $M_1$ in \texttt{view1},

\begin{equation}
    \mathbf{V}_1', \mathbf{M}_1 = \text{GAPW}(\mathbf{V}_2, \mathbf{D}_2; v2\rightarrow v1),
\end{equation}
where $v2\rightarrow v1$ means warping from \texttt{view2} to \texttt{view1}. For the inpainting task, only the target video and its corresponding mask are required as training pairs. Therefore, we add $(\mathbf{V}_1, \mathbf{M}_1)$ to dataset for training the stereo inpainting model.

\begin{figure}[t]
\centering
\includegraphics[width=\columnwidth]{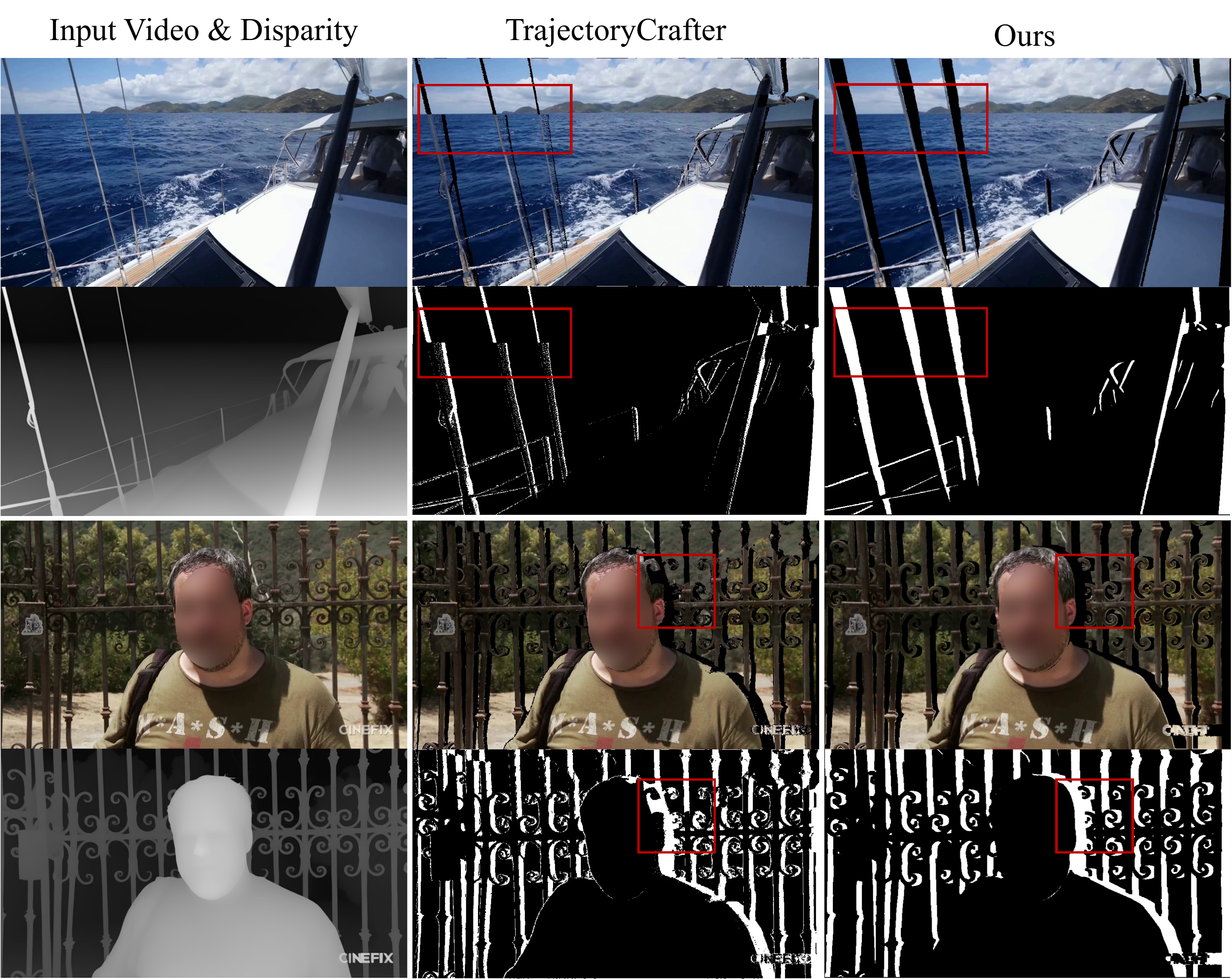} 
\caption{Qualitative comparison of data construction between TrajectoryCrafter~\cite{yu2025trajectorycrafter} and ours.}
\label{fig:datasetdiff}
\vspace{-10pt}
\end{figure}

We compare the construction results between our method and TrajectoryCrafter~\cite{yu2025trajectorycrafter} in Fig.~\ref{fig:datasetdiff}. 
As used in TrajectoryCrafter, forward warping introduces misaligned artifacts in multi-layer background regions, whereas our method leverages GAPW to generate cleaner and geometrically consistent occlusion masks.

\subsection{Sparsity-Aware Stereo Inpainting}
\label{sec:sparsity_aware}

In stereo inpainting task, where the occluded regions to be filled account for only a small portion of pixels, processing all visual tokens equally leads to significant computational redundancy. A straightforward and intuitive approach is to discard part of the visual tokens that do not require inpainting. Our method, developed based on WanVideo~\cite{wan2025}, consists of two key components: a denoiser $ D_\theta $ based on a Diffusion Transformer (DiT), and a 3D Variational Autoencoder (VAE) comprising an Encoder $ \mathcal{E} $ and a Decoder $ \mathcal{D} $, as illustrated in Fig.~\ref{fig：framework} (b). Our primary objective is to perform inpainting on a video $ \mathbf{V}^m \in \mathbb{R}^{3 \times T \times H \times W} $ containing occlusion regions, which are represented by a specialized mask $ \mathbf{M} \in \mathbb{R}^{1 \times T \times H \times W} $. In training, the corrupted video $ \mathbf{V}^m $ is obtained by performing $\mathbf{M}$ on the monocular video $ \mathbf{V} \in \mathbb{R}^{3 \times T \times H \times W} $.

The 3D VAE is used to compress the input videos into a latent space, reducing both temporal and spatial dimensions. Specifically, the Encoder $ \mathcal{E} $ maps both the target video $ \mathbf{V} $ and the corrupted video $ \mathbf{V}^m $ into their respective latents,
\begin{equation}
    \mathbf{z}_0 = \mathcal{E}(\mathbf{V}), \quad \mathbf{z}^m = \mathcal{E}(\mathbf{V}^m),
\end{equation}
where $ \mathbf{z}_0 , \mathbf{z}^m \in \mathbb{R}^{c \times t \times h \times w} $ denote the latent representations of $ \mathbf{V} $ and $ \mathbf{V}^m $, each containing $t \times h \times w$ visual tokens. The mask $\mathbf{M}$ is downsampled to $\mathbf{m} \in \mathbb{R}^{1 \times t \times h \times w}$ to match the shape of $\mathbf{z}^m$. To preserve the spatial and temporal context for inpainting, $\mathbf{z}^m$ undergoes a dilation process and is then used to select visual tokens,

\begin{equation}
    (\mathbf{\hat{z}}_0, \mathbf{\hat{z}}^m, \mathbf{\hat{m}} ) = \mathcal{S}((\mathbf{z}_0, \mathbf{z}^m, \mathbf{m} ), \Phi(\mathbf{m}, k)),
\end{equation}
where $\mathbf{\hat{z}}_0$, $\mathbf{\hat{z}}^m$, $\mathbf{\hat{m}}$ represent the sparse counterparts, $\mathcal{S}(\cdot)$ denotes a mask-based selection function, and $\Phi(\cdot)$ denotes a dilation function with kernel size $k$. Our inpainting process is built on the flow matching framework, which involves a forward noising process and a backward denoising process. During the forward process, noise is added to $ \mathbf{z}_0 $ to simulate a transition from the data distribution to a noise distribution, which can be expressed as,
\vspace{-4pt}
\begin{equation}
\mathbf{\hat{z}}_t = (1 - \sigma_t) \cdot \mathbf{\hat{z}}_0 + \sigma_t \cdot \boldsymbol{\epsilon},
\end{equation}
where $ \sigma_t \in (0, 1] $ controls the noise intensity, and $ \boldsymbol{\epsilon} \sim \mathcal{N}(\mathbf{0}, \mathbf{I}) $ is sampled from the Gaussian distribution. As $ \sigma_t $ increases, the sample $ \mathbf{\hat{z}}_t $ moves to the noise distribution with a constant velocity $ \mathbf{v} = \boldsymbol{\epsilon} - \mathbf{\hat{z}}_0 $.

The denoiser $ D_\theta $ is trained to predict the velocity of $ \mathbf{\hat{z}}_t $ at time $ t $, enabling the reverse denoising process. The training objective is defined as,

\begin{equation}
   \mathcal{L} = \left\| D_{\theta}(\mathbf{\hat{z}}_t, \mathbf{\hat{z}}^m, \mathbf{\hat{m}}, t) - \mathbf{v} \right\|_2    
\end{equation}

To integrate the masked video into the denoiser, we concatenate $ \mathbf{z}_t $, $ \mathbf{\hat{m}} $, and $ \mathbf{\hat{z}}^m $ along the channel dimension. For compatibility, the weights corresponding to these additional input channels are initialized to zero. During inference, the denoiser $  D_\theta  $ performs a multi-step denoising on $  \mathbf{\hat{z}}_t  $  as follows,

\begin{equation}
\mathbf{\hat{z}}_{t-1} = \mathbf{\hat{z}}_t + D_{\theta}(\mathbf{\hat{z}}_t, \mathbf{\hat{z}}^m, \mathbf{\hat{m}}, \mathbf{c}, t) \cdot (\sigma_{t-1} - \sigma_t)    
\end{equation}

After $ N $ inference steps, $ \mathbf{\hat{z}}_0 $ is recovered. Finally, the inpainted video $\mathbf{V}$ can be obtained by

\begin{equation}
 \mathbf{V} = \mathcal{B}(\mathcal{D}(\mathcal{B}(\mathbf{\hat{z}}_0, \mathbf{z}^m, \mathbf{m})), \mathbf{V}^m, \mathbf{M})
\end{equation}
where $\mathcal{B}(\cdot)$ is a mask-based blending function. With a dilation kernel size $k = 3$, only ~25.6\% of the visual tokens participate in the computation, accelerating the inference speed of DiT by $10.7\times$ (see Tab.~\ref{tab:ablation_sparse} for more details).

\begin{figure*}[t]
\centering
\includegraphics[width=\textwidth]{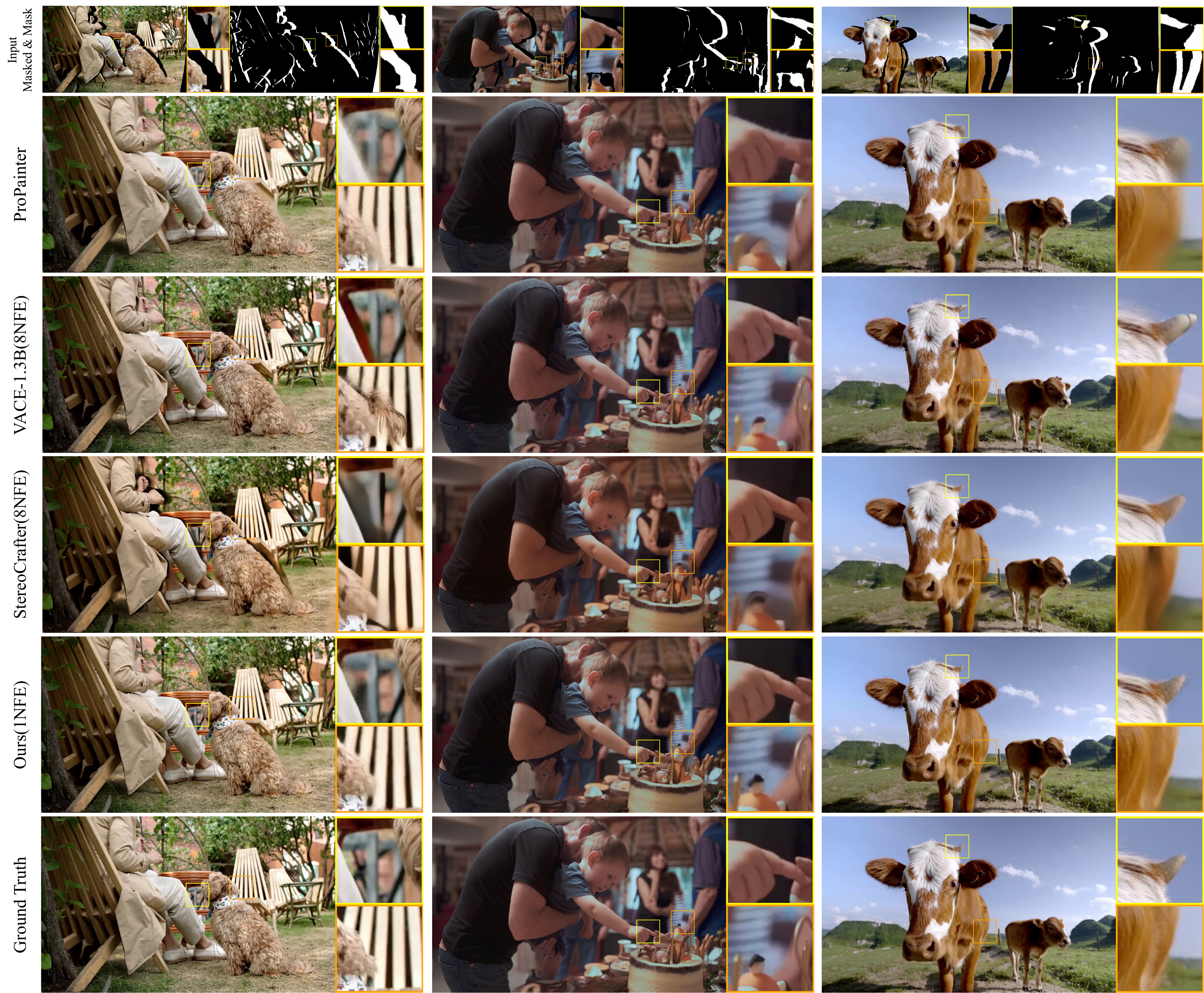}
\caption{Qualitative comparison of stereo video inpainting on the HD-100 test set ($768{\times}1280$). Our inpainting results are much closer to the ground truth, showing sharper details and clearer structures.}
\label{fig:hd100}
\vspace{-12pt}
\end{figure*}

\section{Experiments}

\begin{figure*}[t]
\centering
\includegraphics[width=1.0\textwidth]{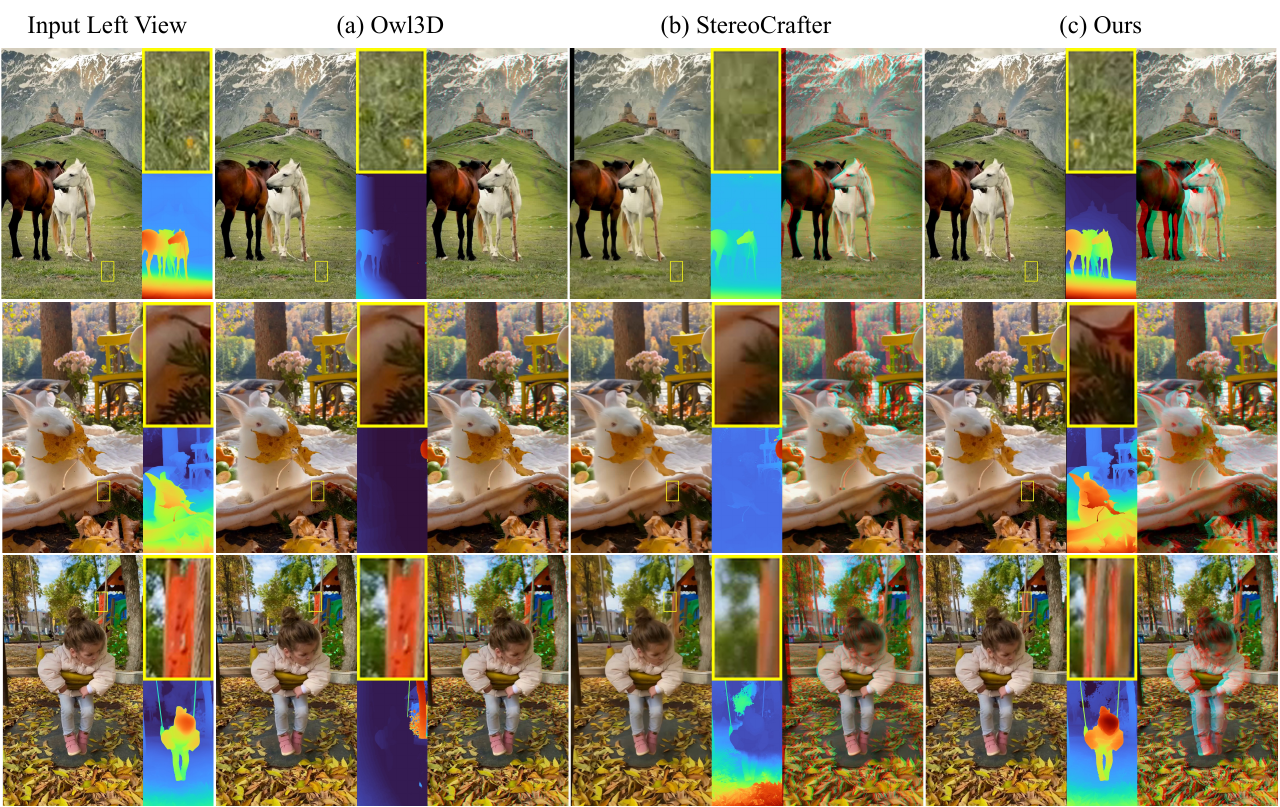}
\caption{Qualitative comparison of 2D-to-3D conversion. Each group shows the \colorbox[HTML]{F5F0FF}{generated right view}, \colorbox[HTML]{FFF7E6}{zoomed-in patches}, \colorbox[HTML]{E6F7FF}{stereo-matched disparity}, and \colorbox[HTML]{E6FFFA}{anaglyph visualization} from left to right. Our method produces sharp, high-quality views with accurate and detailed disparity.}
\label{fig4stereo}
\vspace{-12pt}
\end{figure*}

\subsection{Implementation Details}
\paragraph{Model and Training Scheme.} 
Our model is built upon Wan2.1-1.3B~\cite{wan2025}, whose DiT component is fine-tuned using LoRA adaptation, while a lightweight 3D-aware VAE is distilled separately for faster encoding and decoding following~\cite{zhao2024cvvae} (see supplementary for details).
We adopt a two-stage training scheme on the OpenVid dataset~\cite{nan2024openvid}. 
Stage~1 uses random masks for generic inpainting pretraining. 
Stage~2 employs our proposed method described in Sec .~\ref{sec:pbdp} 
to generate occlusion masks at three target resolutions($1280{\times}720$, $720{\times}1280$, and $768{\times}768$), producing a $56\text{k}{\times}3$ pseudo-stereo inpainting dataset with maximum disparity randomly sampled from $[0.3,0.8]$. 
During training, we adopt fully dense token computation (100\%), while the sparsity-aware strategy is explored and applied only at inference time for acceleration.
Training is conducted on 4 NVIDIA A100 GPUs with a learning rate of $4\times10^{-5}$. 
Stage~1 and Stage~2 are trained for 10k and 2.5k steps with batch sizes of 12 and 2, respectively.
\paragraph{Evaluation Datasets and Metrics.} 
We evaluate on three stereo test sets with unified 81-frame clips. 
\textbf{HD-100} is a real-world dataset collected from internet 4K videos (100 samples), where the inpainting ground truth is the original RGB sequence and masks are generated following the same procedure as in Stage~2.
\textbf{Dynamic Replica}~\cite{karaev2023dynamicstereo} is a synthetic stereo dataset (20 validation samples) with ground-truth disparity, while 
\textbf{SVD}~\cite{izadimehr2025svd} is a spatial video dataset (20 samples) randomly selected from the Apple Vision Pro (AVP) subset (baseline 63.76\,mm), with disparity estimated via a stereo matching method~\cite{wen2025stereo}. 
For both stereo datasets, the left view is used as input and the right view as ground truth, and the corresponding masks and warped inputs are constructed using the method described in Sec.~\ref{sec:gapw}.
We report PSNR~\cite{wikipedia2024psnr}, SSIM~\cite{wang2004image}, and LPIPS~\cite{zhang2018unreasonable} as evaluation metrics.

\begin{table}[t]
\small
\centering
\setlength{\tabcolsep}{3pt}
\begin{tabular}{lccccc}
\toprule
\textbf{Method} & \textbf{NFE} & \textbf{Latency}$\downarrow$ & \textbf{PSNR}$\uparrow$ & \textbf{SSIM}$\uparrow$ & \textbf{LPIPS}$\downarrow$  \\
\midrule
ProPainter$^{\dagger}$ & 1 & 668.1 & 28.30 & 0.927 & 0.052  \\
VACE\mbox{-}1.3B & 8  & 2779.1   & 25.38 & 0.859 & 0.095     \\
ZeroStereo  &  8  & 506.0      & 19.12  & 0.676 & 0.189    \\
ZeroStereo & 50    & 2024.7     & 24.73  & 0.771 & 0.094     \\
StereoCrafter & 8  & 716.5      & 23.99 & 0.782 & 0.142 \\
\textbf{Ours} & 1  & 40.1     & 30.48 & 0.900 & 0.053  \\
\textbf{Ours (blended)} & 1 & \textbf{40.1} & \textbf{32.65} & \textbf{0.948} & \textbf{0.026}  \\
\bottomrule
\end{tabular}
\caption{\textbf{HD-100 @ $768{\times}1280$}. Quality and latency (ms/frame). 
$^{\dagger}$ non-diffusion.}
\label{tab:hd100}
\vspace{-12pt}
\end{table}

\subsection{Comparison of Stereo Inpainting}
We evaluate on the HD-100 benchmark using unified inputs and timing setup on an NVIDIA A100 for fairness. 
Tab.~\ref{tab:hd100} presents quantitative comparisons in terms of PSNR, SSIM, and LPIPS across both stereo-specific methods (StereoCrafter, ZeroStereo) and general video inpainting models (VACE-1.3B~\cite{jiang2025vace}, ProPainter~\cite{zhou2023propainter}).
For brevity, we show three representative methods in the main paper; complete results for all four baselines are provided in the appendix, along with visual comparisons to two non-released methods~\cite{2024spatialdreamer,shvetsova2025m2svid}.
Our approach achieves the best accuracy while maintaining real-time inference.
Specifically, ProPainter achieves the second-best performance but produces noticeably blurry outputs (see Fig.~\ref{fig:hd100}), while ZeroStereo shows a strong dependence on diffusion steps, and its performance drops sharply when the default NFE=50 is reduced to 8.
In contrast, our model performs \emph{single-step} inference (NFE=1) in real time (40.1ms $\approx$ 25fps) while preserving fine stereo structures.
Furthermore, we introduce a \emph{blended} variant that leverages GAPW-derived masks for geometry-aligned region fusion. Unlike previous methods using forward warping, such as StereoCrafter, whose masks exhibit scattered edge artifacts, our masks are cleaner and geometrically consistent, enabling stable blending at the native input resolution and yielding further perceptual gains.

\begin{table*}[t]
\small
\centering
\setlength{\tabcolsep}{15pt}
\begin{tabular}{lccccccc}
\toprule
& & \multicolumn{3}{c}{\textbf{Dynamic Replica }} & \multicolumn{3}{c}{\textbf{SVD}} \\
\cmidrule(lr){3-5} \cmidrule(lr){6-8}
\textbf{Method} & \textbf{NFE} &  \textbf{PSNR}$\uparrow$ & \textbf{SSIM}$\uparrow$ & \textbf{LPIPS}$\downarrow$
                & \textbf{PSNR}$\uparrow$ & \textbf{SSIM}$\uparrow$ & \textbf{LPIPS}$\downarrow$ \\
\midrule
Deep3D$^{\dagger}$   & 1       & 17.11 & 0.711 & 0.252 & 19.32 & 0.590 & 0.224 \\
Depthify.ai$^{\dagger}$ & 1    & 18.29 & 0.727 & 0.191 & 20.70 & 0.617 & 0.190 \\
Owl3D$^{\dagger}$  & 1         & 17.87 & 0.726 & 0.209 & 19.99 & 0.615 & 0.214 \\
StereoCrafter & 8  & 17.93 & 0.737 & 0.226 & 24.68 & 0.752 & 0.199 \\
ZeroStereo & 50  & 23.08 & 0.872 & 0.105
                & 23.81 & 0.725 & 0.148 \\
\textbf{Ours}  & 1 & 29.12 & 0.920 & 0.063
                & 24.88 & 0.776 & 0.131 \\
\textbf{Ours (blended)} & 1 & \textbf{29.17} & \textbf{0.922} & \textbf{0.057}
                & \textbf{25.30} & \textbf{0.793} & \textbf{0.113} \\
\bottomrule
\end{tabular}
\caption{Quantitative comparison of 2D-to-3D methods on the Dynamic Replica valid set and the SVD AVP test set.$^{\dagger}$End-to-end approaches.}
\label{tab:replica_dual}
\vspace{-12pt}
\end{table*}

\subsection{Comparison of 2D-to-3D Conversion}
To further assess cross-domain generalization, we evaluate our method on two stereo datasets: the Dynamic Replica synthetic set and the SVD real-world stereo video dataset collected via AVP.  
Both datasets provide ground-truth right views, enabling quantitative evaluation of 2D-to-3D conversion accuracy.  
As summarized in Tab.~\ref{tab:replica_dual}, our method consistently achieves the best PSNR, SSIM, and LPIPS across both datasets, outperforming end-to-end approaches~\cite{xie2016deep3d, owl3d, depthify} and warp-and-inpaint baselines~\cite{zhao2024stereocrafter, wang2025zerostereo}.  

For clearer visualization, Fig.~\ref{fig4stereo} presents 2D-to-3D conversion results on HD-100, where no ground-truth right views are available. Accordingly, this figure is intended for visual comparison only, while additional results on Replica and SVD are provided in the appendix for completeness.
The comparison includes our method, the industrial solution Owl3D, and the academic baseline StereoCrafter, all evaluated using their released default settings.
In both the \emph{horse} and \emph{rabbit} scenes, our model reconstructs sharper textures and cleaner, well-aligned disparity, delivering a more coherent and immersive 3D experience.

\vspace{-4pt}

\subsection{Ablation Study} 
\begin{table}
\small
\centering
\setlength{\tabcolsep}{10pt}
\begin{tabular}{lccc} 
\toprule
\textbf{Dataset Strategy} & \textbf{PSNR} $\uparrow$  & \textbf{SSIM} $\uparrow$  & \textbf{LPIPS} $\downarrow$ \\ 
\hline
Random Mask                          & 26.64          & 0.906          & 0.092           \\
TrajectoryCrafter                    & 31.14          & 0.923          & \textbf{0.047}  \\
\textbf{Ours}                        & \textbf{32.48} & \textbf{0.933} & 0.049           \\
\bottomrule
\end{tabular}
\caption{Ablation of inpainting performance on dataset strategy.}
\label{tab:ablation_dataset}
\vspace{-12pt}
\end{table}
\paragraph{Data Construction Matters.}
We first assess the impact of our pseudo-stereo data pipeline. Compared with TrajectoryCrafter (Fig.~\ref{fig:datasetdiff}), our construction yields inpainting pairs with stronger geometric consistency and cleaner occlusion masks, especially around fine structures and distant backgrounds. As shown in Tab.~\ref{tab:ablation_dataset}, fine-tuning on our dataset achieves the best PSNR/SSIM on HD-100 for both the Stage-1 baseline and the TrajectoryCrafter-trained counterpart, indicating that more accurate pseudo-stereo construction directly improves downstream stereo inpainting. To isolate the effect of data construction, this ablation keeps the architecture and inference configuration identical across rows (dense inference, original VAE, \(576{\times}1024\) resolution, NFE=8), while the final model reported elsewhere incorporates these orthogonal improvements.

\vspace{-10pt}

\begin{table}[t]
\small
\centering
\setlength{\tabcolsep}{2pt}
\begin{tabular}{llccccc}
\toprule
\textbf{Dilate} & \textbf{Stride} & \textbf{$r$ (\%)} & \textbf{Latency$^{\ddagger}$}$\downarrow$ & \textbf{PSNR}$\uparrow$ & \textbf{SSIM}$\uparrow$ & \textbf{LPIPS}$\downarrow$ \\
\midrule
\multicolumn{2}{l}{\textit{baseline}} & 100.0 & 380.9 & \textbf{32.48} & \textbf{0.933} & \textbf{0.049} \\
\midrule
— & — & 13.7 & \textbf{18.51} & 29.84 & 0.916 & 0.063 \\
3 & — & 25.6 & 35.7 & 32.01 & 0.931 & 0.051 \\
3 & 20   & 29.4 & 44.8 & 32.10 & 0.931 & \underline{0.050} \\
3 & 10   & 33.2 & 54.8 & 32.16 & 0.931 & \underline{0.050} \\
\rowcolor{gray!15}
5 & — & 36.1 & 58.4 & 32.31 & \underline{0.932} & \underline{0.050} \\
5 & 20   & 39.4 & 68.6 & 32.34 & \underline{0.932} & \underline{0.050} \\
5 & 10   & 42.7 & 78.2 & \underline{32.36} & \underline{0.932} & \underline{0.050} \\

\bottomrule
\end{tabular}
\caption{HD-100 ($576{\times}1024$) ablation with fixed NFE$=4$, varying dilation kernel and temporal stride. 
Retention rate $r$ is the fraction of tokens kept for DiT computation; $^{\ddagger}$Latency denotes DiT-only inference time in ms/frame. 
All settings achieve a substantial speedup with negligible quality loss. 
The light-gray row (\textit{dilate=5}) indicates the optimal setting adopted in all other experiments.}
\label{tab:ablation_sparse}
\vspace{-10pt}
\end{table}

\paragraph{Sparsity-Aware Acceleration with Negligible Quality Loss.}
Building on Sec .~\ref{sec:sparsity_aware}, we evaluate our \emph{sparsity-aware} design at fixed NFE by varying the token retention rate \(r\) guided by GAPW masks. 
We adjust the retention rate \(r\) by spatially dilating the GAPW mask and temporally striding dense tokens, keeping only those essential for stereo completion, namely the tokens within the occlusion mask \(\mathbf{m}\) and a narrow dilated band around its boundary, while pruning the rest from DiT computation.
As shown in Tab.~\ref{tab:ablation_sparse}, retention rates from 43\% to 14\% significantly reduce DiT inference latency with only negligible quality changes: SSIM/LPIPS remain stable and PSNR stays comparably high on HD-100. 
Thus, GAPW-guided, structure-aware token selection effectively removes redundant computation while preserving geometric fidelity and perceptual quality. 
Balancing efficiency and robustness, we adopt \(r\!\approx\!35\%\) (e.g., dilation kernel \(=5\)) as the default.

\vspace{-5pt}

\begin{figure}[ht]
\centering
\includegraphics[width=\columnwidth]{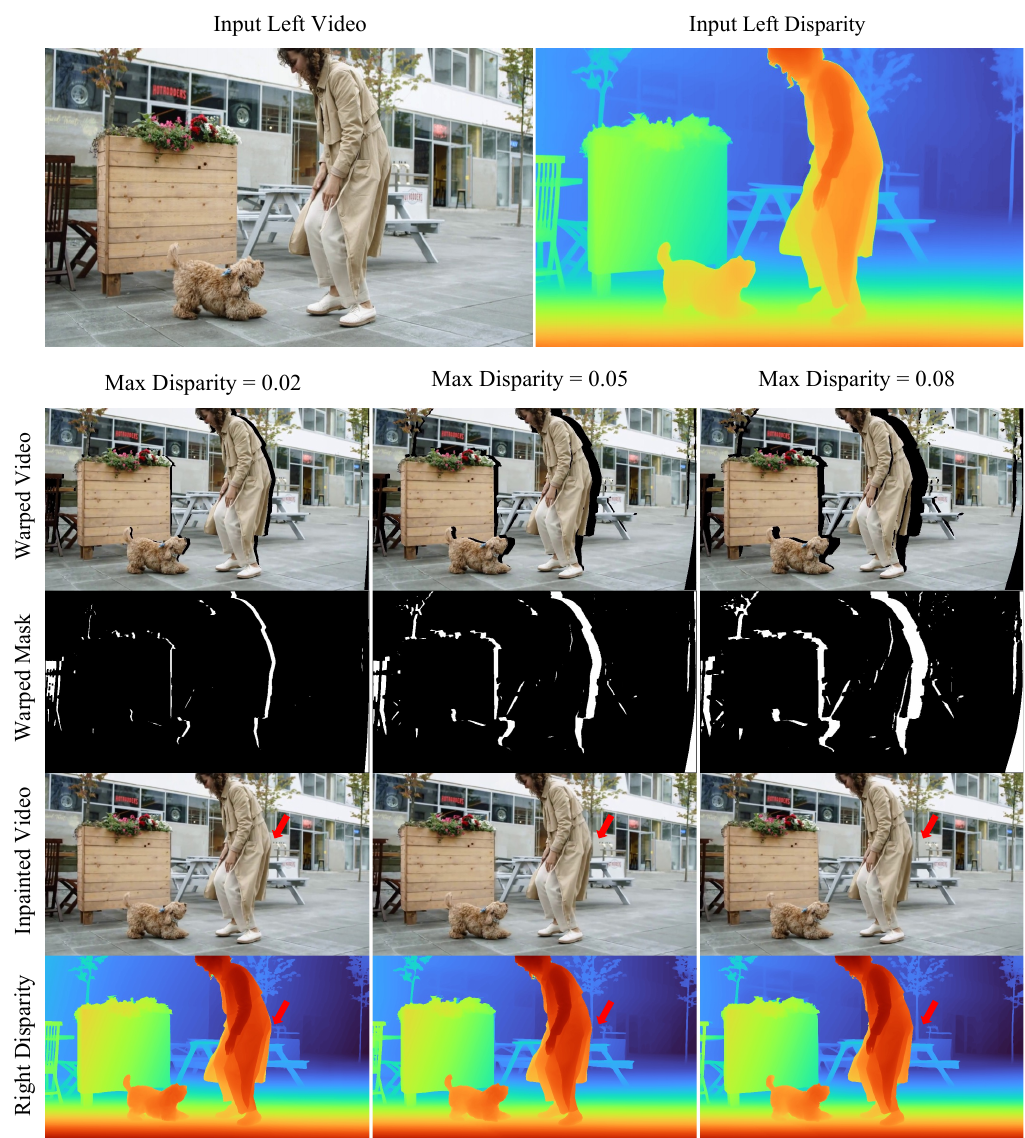} 
\caption{Ablation on max disparity. Our method consistently delivers high-quality, geometrically accurate results even under large disparity ranges.}
\label{fig:ablation_maxdisp}
\vspace{-15pt}
\end{figure}

\begin{table}[t]
\small
\centering
\setlength{\tabcolsep}{11pt}
\begin{tabular}{cccc}
\toprule
\textbf{Max Disparity} & \textbf{PSNR}$\uparrow$ & \textbf{SSIM}$\uparrow$ & \textbf{LPIPS}$\downarrow$ \\
\midrule
0.02 & \textbf{32.18} & \textbf{0.915} & \textbf{0.043} \\
0.05 & 32.16 & 0.905 & 0.049 \\
\rowcolor{gray!15}  
0.07 & 30.48 & 0.900 & 0.053 \\
0.08 & 29.91 & 0.895 & 0.058 \\
\bottomrule
\end{tabular}
\caption{HD-100 ($768{\times}1280$) ablation with fixed NFE$=1$ under wide-baseline settings. 
The light-gray row ($0.07$) corresponds to the setting adopted for all stereo inpainting experiments on the HD-100 test set. 
Larger disparities enlarge occluded areas, causing a slight expected drop in metrics, yet our method remains robust and effective.}
\label{tab:ablation_wide}
\vspace{-10pt}
\end{table}

\vspace{-5pt}

\paragraph{Wide-Baseline Robustness.}
Finally, we evaluate robustness under increasing maximum disparity, which simulates wider stereo baselines, analogous to simulating human interocular distance and introducing more occluded regions to inpaint. 
Figure~\ref{fig:ablation_maxdisp} visualizes the inpainting results and right-view disparity estimates under three disparity settings, showing that our method consistently preserves geometry and visual quality as the baseline widens. 
Quantitative results in Tab.~\ref{tab:ablation_wide} further confirm that even as occluded regions expand, our single-step model (NFE=1) remains robust with only minor metric drops. 
In contrast, Fig.~\ref{fig4stereo} shows that Owl3D and StereoCrafter degrade even at smaller disparities. 
Overall, our approach demonstrates strong resilience to large-baseline stereo, enabling stereo video generation with human-eye scale separation for an immersive 3D viewing experience.

\vspace{-5pt}

\section{Conclusion}

This paper is dedicated to the task of 2D-to-3D stereo video inpainting. To address the issues of scattered pixels and discontinuous masks in warped videos caused by the commonly used forward warping, we propose Gradient-Aware Parallax Warping (GAPW), which combines backward warping and the gradients of the coordinate mapping function to obtain smoother and more continuous edges. Subsequently, based on the proposed GAPW, we introduce Parallax-Based Dual Projection (PBDP), which obtains the occlusion mask of the input view through two projections to construct stereo inpainting data without the need for stereo videos. This high-quality data construction method enables our approach to achieve state-of-the-art performance on the stereo inpainting task. Finally, we present Sparsity-Aware Stereo Inpainting (SASI) to reduce the computational redundancy in such an inpainting task. By discarding more than 70\% redundant tokens, SASI achieves a 10.7× speedup in DiT inference, while the performance loss is almost negligible compared to its dense computation counterpart, enabling us to process HD videos at real-time speed on a single A100 GPU.

\clearpage
\setcounter{page}{1}
\maketitlesupplementary
\appendix

\section*{Appendix}
\noindent
This supplementary material provides additional analyses and visual results that complement the experiments presented in the main paper. 
Appendix~A details the runtime evaluation and component ablations, while Appendix~B presents extended qualitative comparisons across datasets and baselines, together with representative failure cases.

\section{Extended Experiments}

\subsection{Runtime Analysis}
\label{sec:part_speed}
\begin{figure}[t]
\centering
\includegraphics[width=\columnwidth]{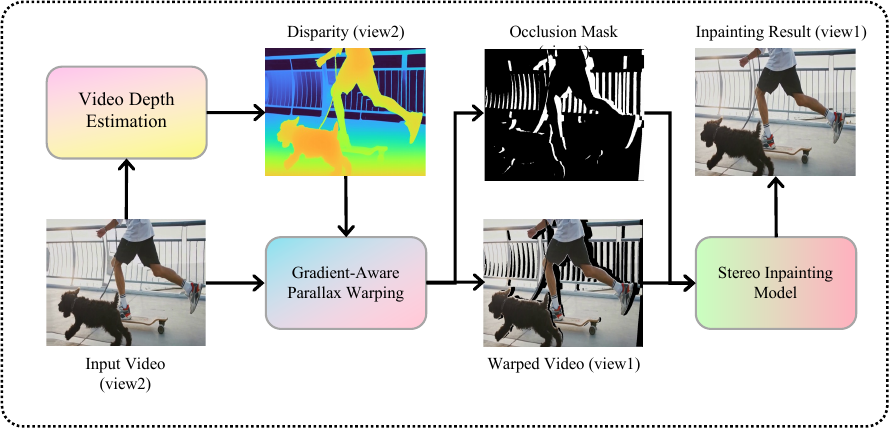}
\caption{Inference pipeline of 2D-to-3D conversion.}
\label{fig:supp_dreamstereo}
\end{figure}

Figure~\ref{fig:supp_dreamstereo} illustrates the inference pipeline of 2D-to-3D conversion, which integrates our proposed Gradient-Aware Parallax Warping (GAPW) for precise occlusion handling and view synthesis. 
Within this pipeline, DreamStereo serves as the stereo inpainting module responsible for reconstructing occluded regions in the right view. 
We further analyze the runtime characteristics of this module under identical inference configurations.

\begin{figure}[t]
\centering
\includegraphics[width=\columnwidth]{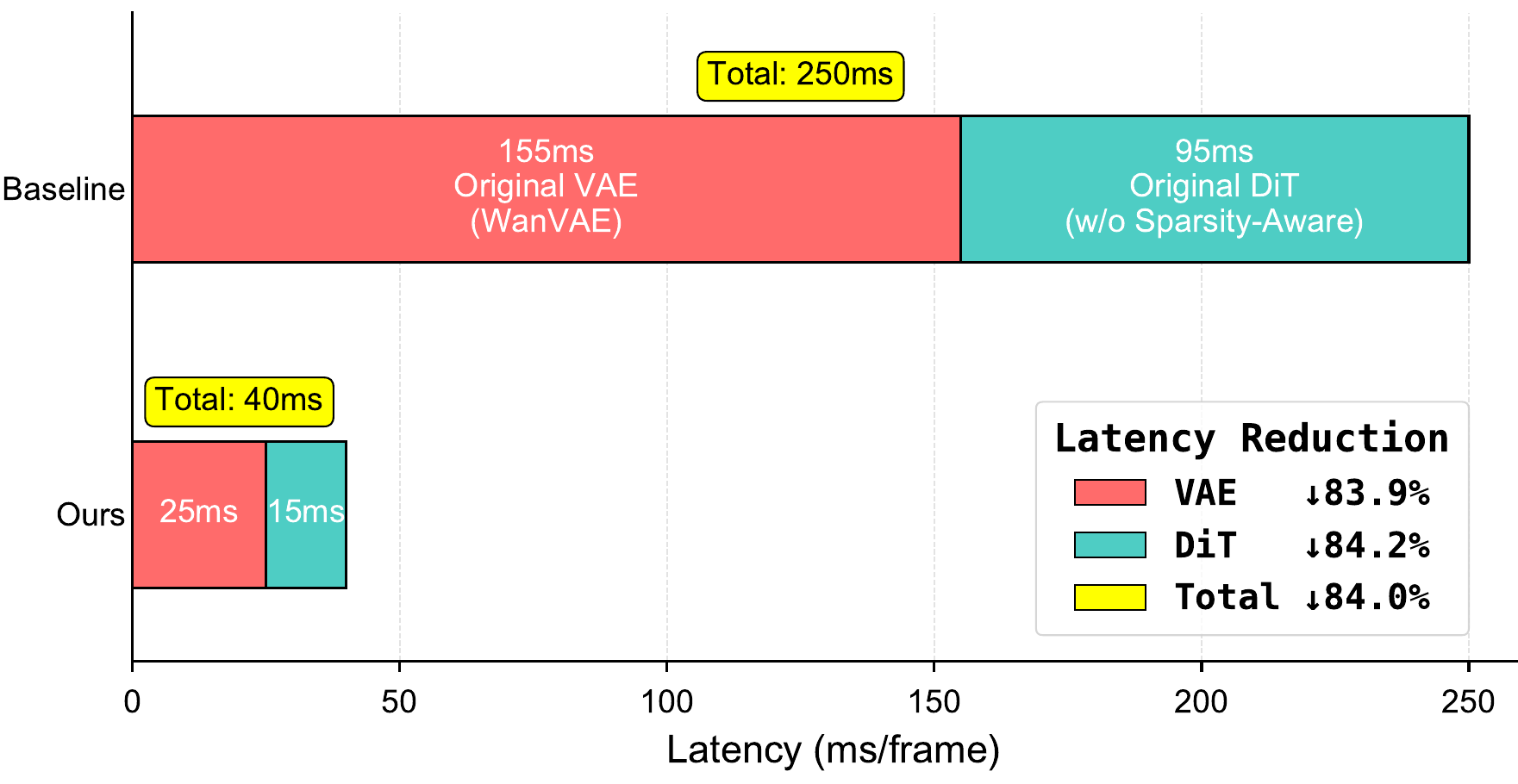}
\caption{Module-wise latency breakdown and reduction on $768{\times}1280$ HD videos.}
\label{fig:part_speed}
\end{figure}

We compare the end-to-end latency of SASI with our internal baseline that adopts dense tokens (no sparsity) and the original WanVAE. 
All measurements were conducted on an NVIDIA A100 using HD-100 videos at $768{\times}1280$, batch size 1, and FP16 precision. 
As shown in Fig.~\ref{fig:part_speed}, our method achieves an overall latency reduction of \textbf{84.0\%} per frame. 
Specifically, the sparsity-aware DiT reduces computation time by \textbf{84.2\%}, while the distilled 3D-aware VAE lowers encoding and decoding cost by \textbf{83.9\%}. 
Together, these two optimizations shorten total inference from 250 to 40 ms per frame, enabling real-time HD stereo generation at 25 fps. 
We also report throughput on commodity GPUs, achieving \textbf{54.2} ms/frame on an RTX 3090 and \textbf{33.0} ms/frame on an RTX 4090, compared with \textbf{40.1} ms/frame on an A100.

\subsection{Distilled VAE Ablation}
\label{sec:slimvae}
We adopt a method similar to CV-VAE~\cite{zhao2024cvvae} to distill the WanVideo~\cite{wan2025} VAE, aiming to reduce the time consumption of video encoding and decoding.
Tab.~\ref{tab:vae_ab_psnr} reports stereo inpainting results on the HD-100 test set using different VAE while keeping the same final setting as in the main paper, showing that the distilled VAE achieves nearly identical quality to the original WanVAE.
For standalone VAE profiling at $1024{\times}1024$ resolution (Tab.~\ref{tab:vae_ab_speed}), our model reduces parameter count by \textbf{36\%} and achieves over $4\times$ faster encoding and decoding, providing substantial acceleration with negligible quality impact.

\begin{table}[h]
\small
\centering
\setlength{\tabcolsep}{12pt}
\begin{tabular}{lccc} 
\toprule
\textbf{VAE} & \textbf{PSNR} $\uparrow$  & \textbf{SSIM} $\uparrow$  & \textbf{LPIPS} $\downarrow$ \\ 
\midrule
WanVAE   & 30.59  & 0.906   & 0.040 \\
Distilled (Ours) & 30.48  & 0.900   & 0.053 \\
\bottomrule
\end{tabular}
\caption{Ablation on stereo inpainting quality on HD-100 ($768{\times}1280$).
Results are evaluated under the same final setting as in the main paper (see Tab.~\ref{tab:hd100}), using different VAEs.
The distilled VAE achieves nearly identical quality to the original WanVAE.}
\label{tab:vae_ab_psnr}
\end{table}

\begin{table}[h]
\small
\centering
\setlength{\tabcolsep}{6pt}
\begin{tabular}{lcccc} 
\toprule
\textbf{VAE} & \textbf{Params}  & \textbf{Encoder}  & \textbf{Decoder} & \textbf{Total} \\ 
\midrule
WanVAE   & 126.8\,M  & 47.7\,ms   & 75.0\,ms  & 122.7\,ms \\
Distilled (Ours) & 80.2\,M   &  9.3\,ms   & 18.5\,ms  &  27.8\,ms \\
\bottomrule
\end{tabular}
\caption{Ablation on VAE parameter efficiency and latency at $1024{\times}1024$ resolution.}
\label{tab:vae_ab_speed}
\end{table}

\subsection{Training Sparsity Ablation}
\label{sec:train_sparse_ab}
We further conduct an ablation on applying sparsity during training. Tab.~\ref{tab:train_sparse_ab} compares the default setting, which uses sparsity only at inference, with a variant that applies sparsity in both training and inference. The results show that introducing sparsity during training brings no clear performance gain across metrics. Therefore, we use sparsity only at inference in the final model, which reduces redundant computation without changing the training pipeline.

\begin{table}[h]
\small
\centering
\setlength{\tabcolsep}{8pt}
\begin{tabular}{lccc}
\toprule
\textbf{Train-time token filtering} & \textbf{PSNR}$\uparrow$ & \textbf{SSIM}$\uparrow$ & \textbf{LPIPS}$\downarrow$ \\
\midrule
No (dense training)  & 32.48 & 0.933 & 0.049 \\
Yes (sparse training) & 32.31 & 0.932 & 0.050 \\
\bottomrule
\end{tabular}
\caption{Ablation on applying sparsity during training.}
\label{tab:train_sparse_ab}
\end{table}

\begin{figure*}[t]
\centering
\includegraphics[width=\textwidth]{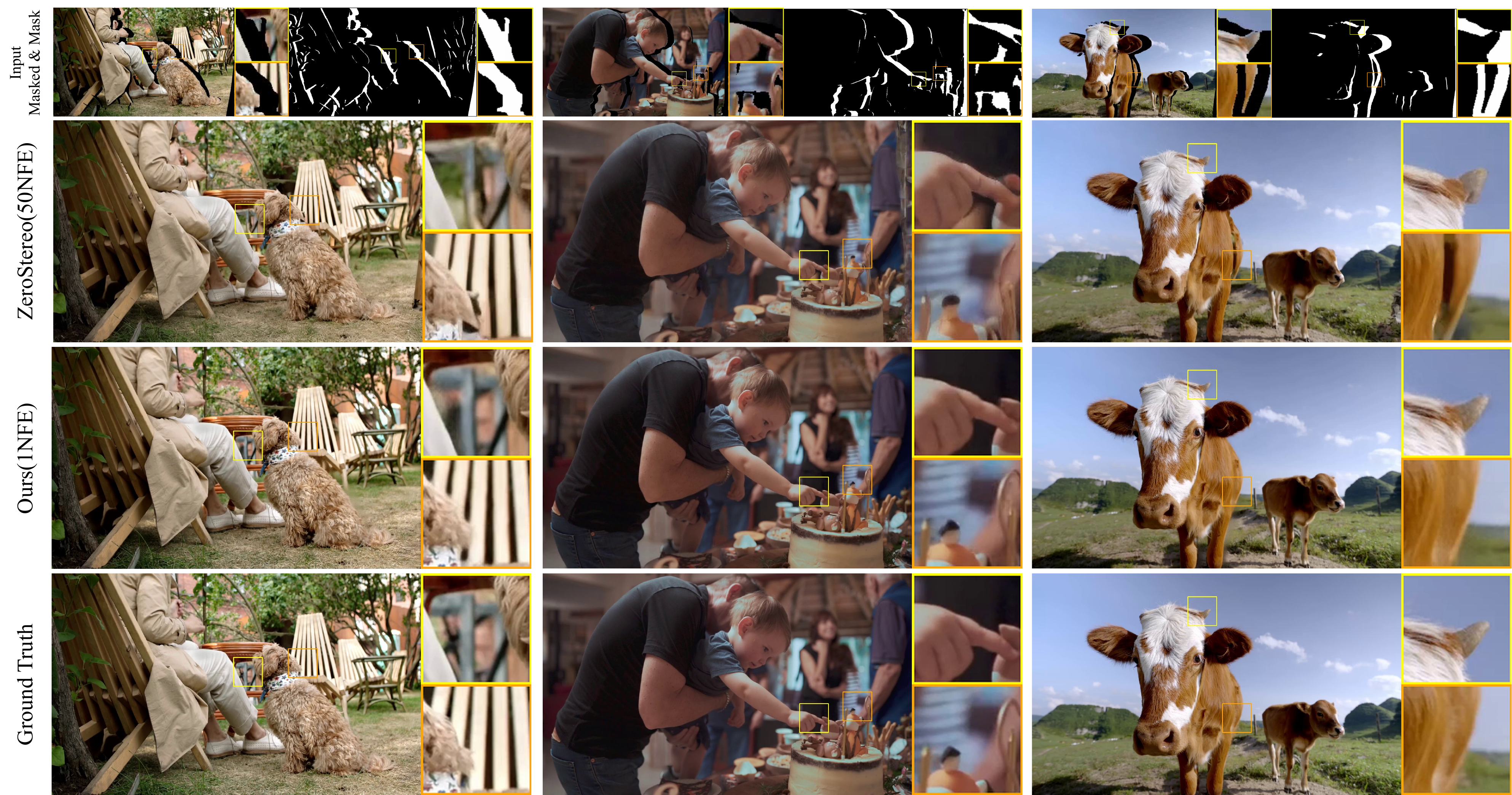}
\caption{\textbf{Qualitative comparison on HD-100 ($768{\times}1280$).}
This figure complements Fig.~\ref{fig:hd100} by including ZeroStereo~\cite{wang2025zerostereo}, a training-free stereo inpainting method that often produces random, spatially inconsistent fillings and distorted textures.}
\label{fig:hd100_supp}
\end{figure*}

\section{Additional Visual Results}
\label{sec:add_vis}

\subsection{Comparison with ZeroStereo}
\label{sec:zerostereo}

Figure~\ref{fig:hd100_supp} complements Fig.~\ref{fig:hd100} in the main paper by adding the stereo inpainting baseline ZeroStereo~\cite{wang2025zerostereo}.
ZeroStereo is a training-free, single-image stereo inpainting method that fails to maintain temporal coherence and often yields distorted textures and geometry.
In contrast, our approach produces temporally consistent, geometrically accurate, and perceptually clean stereo reconstructions.

\subsection{Comparison with SpatialDreamer and M2SViD}
\begin{figure*}[t]
\centering
\includegraphics[width=\textwidth]{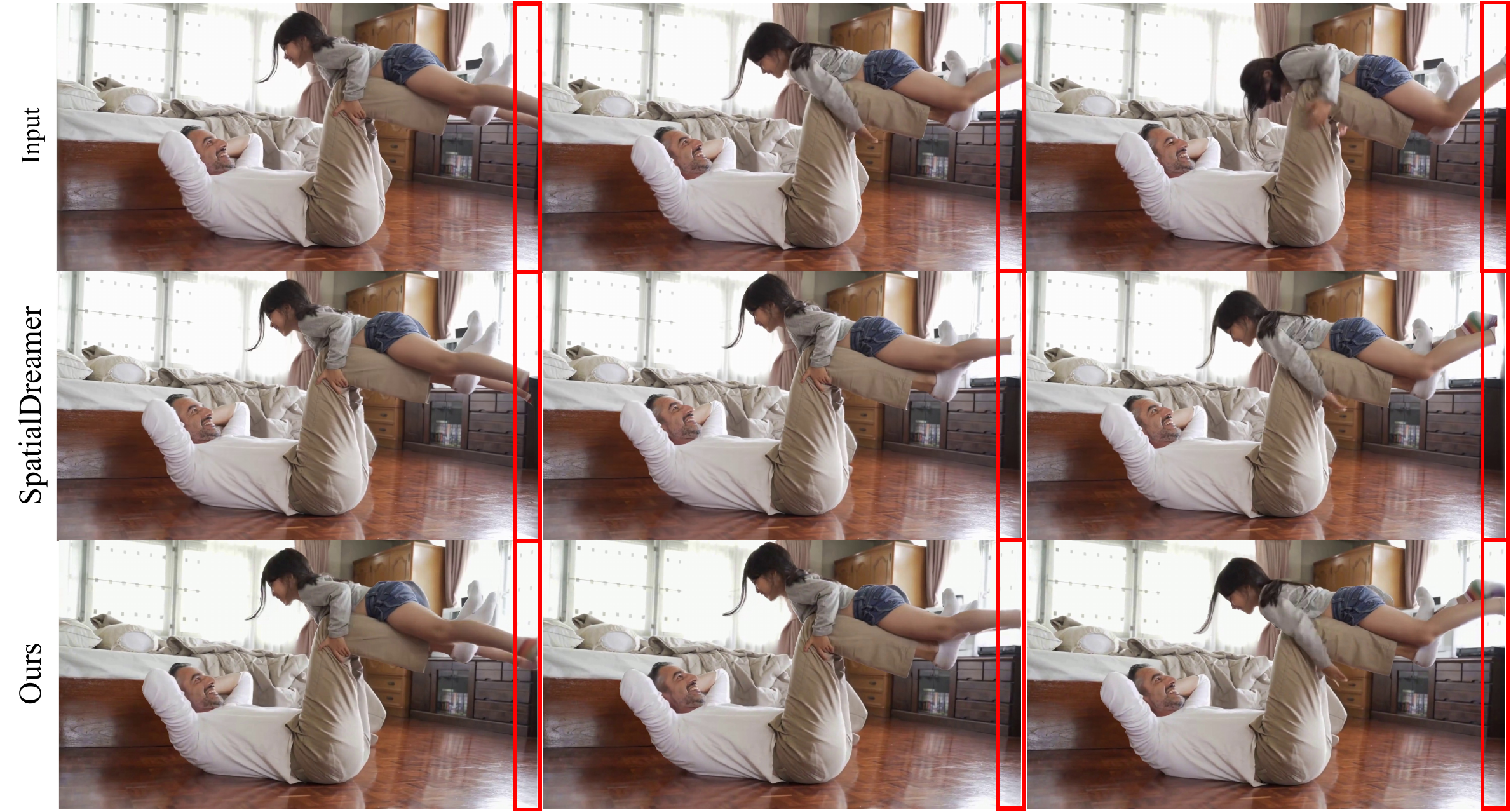}
\caption{\textbf{Comparison with SpatialDreamer~\cite{2024spatialdreamer}.}
Highlighted regions show truncation artifacts in SpatialDreamer, whereas our results preserve complete and consistent stereo geometry.}
\label{fig:spatial_supp}
\end{figure*}

\begin{figure*}[t]
\centering
\includegraphics[width=\textwidth]{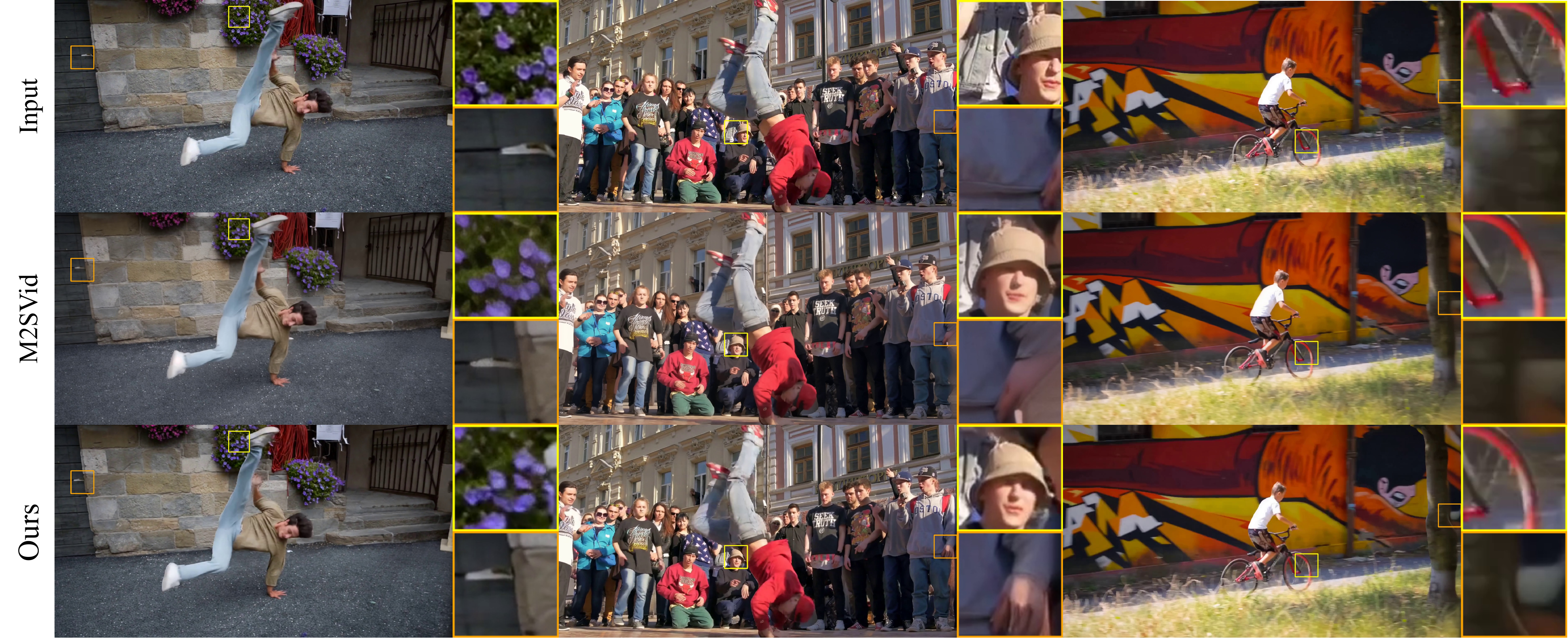}
\caption{\textbf{Comparison with M2SViD~\cite{shvetsova2025m2svid}.}
Our results show sharper details, larger disparities, and better color consistency, while M2SViD suffers from color drift and structural degradation.}
\label{fig:m2svid_supp}
\end{figure*}

We further compare our results with recent non-released stereo synthesis models using publicly available demo videos.
The left view is used as input, and the generated right view is compared.
As illustrated in Figs.~\ref{fig:spatial_supp} and~\ref{fig:m2svid_supp},
SpatialDreamer~\cite{2024spatialdreamer} exhibits truncation artifacts near image boundaries,
while M2SViD~\cite{shvetsova2025m2svid} shows color shifts and degraded structure fidelity.
Our approach maintains complete, sharp, and geometrically aligned stereo structures.

\subsection{Qualitative Results on Dynamic Replica and SVD}
\label{sec:replica_supp}
\begin{figure*}[t]
\centering
\includegraphics[width=0.9\textwidth]{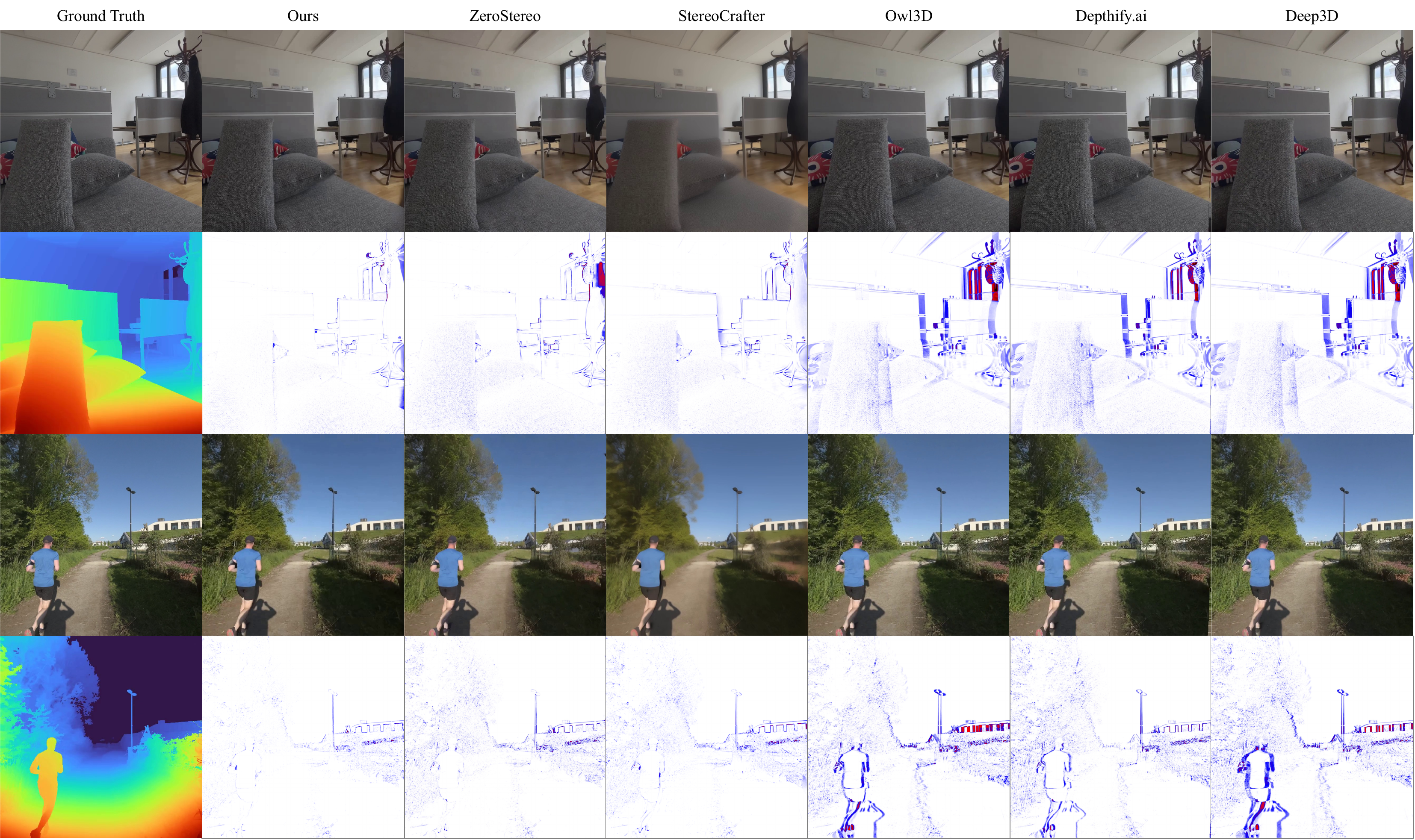}
\caption{\textbf{Qualitative comparison on the SVD AVP test set ($768{\times}768$).}
The second row shows per-pixel MSE maps, where blue denotes small errors and red large deviations from the ground truth. 
Our results exhibit the smallest errors and best overall consistency.}
\label{fig:svd_supp}
\end{figure*}

While the main paper reports quantitative results on SVD AVP and Dynamic Replica test sets,
we provide qualitative comparisons in Figs.~\ref{fig:svd_supp} and ~\ref{fig:replica_supp}.
Our approach produces faithful stereo completions with sharper details and fewer artifacts,
consistent with the quantitative improvements observed in the main paper.

\noindent
\begin{table}
\centering
\scriptsize
\setlength{\tabcolsep}{2pt}
\begin{tabular}{lcccccc}
\toprule
\textbf{Metrics} & \textbf{Deep3D} & \textbf{Depthify.ai}& \textbf{Owl3D} & \textbf{StereoCrafter} & \textbf{ZeroStereo} & \textbf{Ours} \\
\midrule
AbsRel. $\downarrow$ & 0.029 & 0.035 & 0.042 & 0.039 & \textbf{0.018} & \underline{0.019} \\
$\delta<1.05$ $\uparrow$& 0.847 & 0.770 & 0.716 & 0.751 & \textbf{0.929} & \underline{0.914}  \\
\bottomrule
\end{tabular}
\caption{Geometry-consistency comparison on Dynamic Replica.}
\label{tab:depth_ab}
\end{table}

\subsection{Geometry-Consistency Comparison}
\label{sec:zerostereo_metric}
\begin{figure*}[t]
\centering
\includegraphics[width=\textwidth]{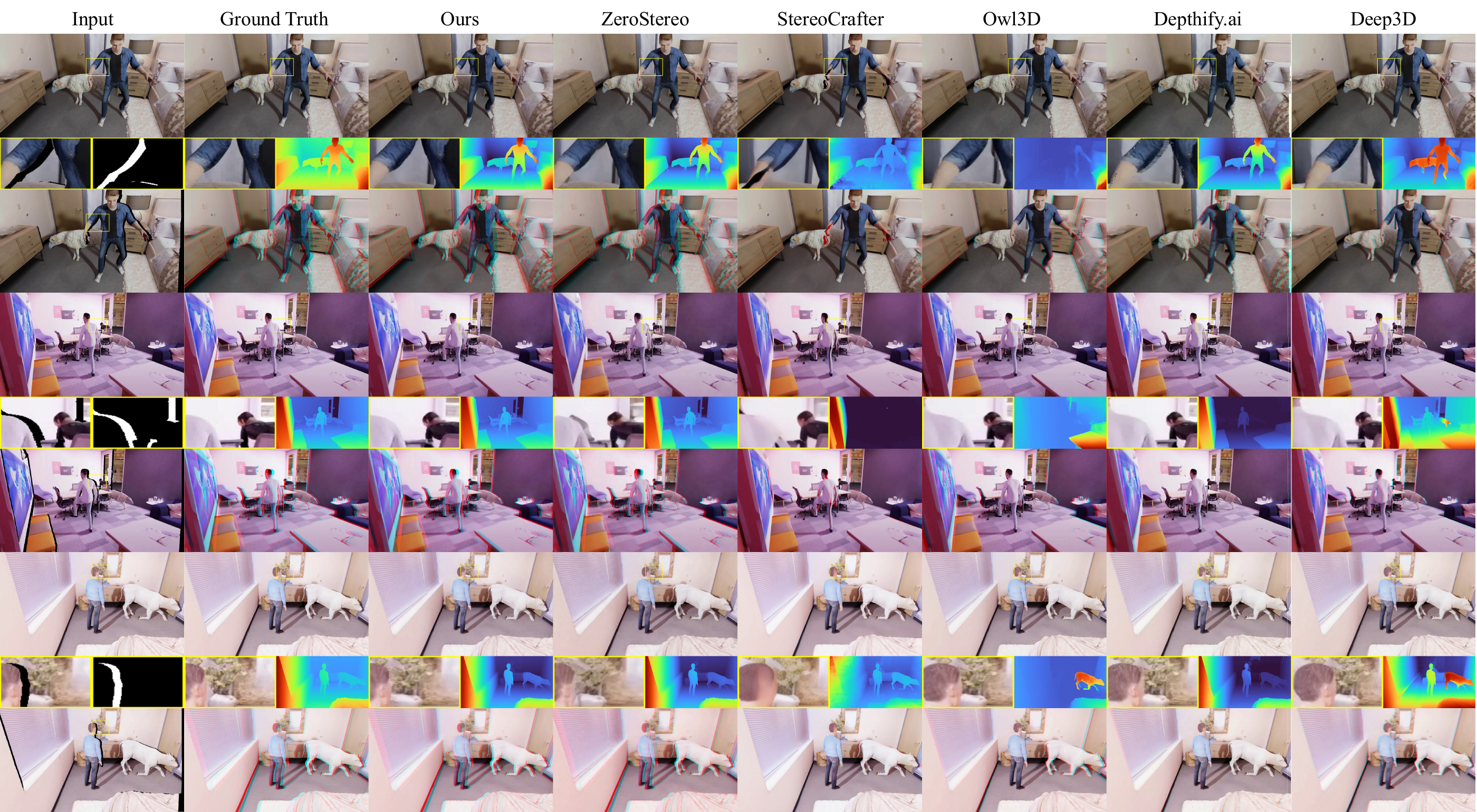}
\caption{\textbf{Qualitative comparison on the Dynamic Replica valid set($720{\times}1280$).}
Our method yields the most accurate and artifact-free stereo completions, consistent with quantitative improvements in Tab.~\ref{tab:replica_dual}.}
\label{fig:replica_supp}
\end{figure*}

We therefore add a geometry-consistency metric on \textbf{Dynamic Replica} (Table~\ref{tab:replica_dual}), where disparity is estimated from the generated left-right pair and depth accuracy is evaluated after alignment to metric depth using scale and shift (Table~\ref{tab:depth_ab}). 
While ZeroStereo achieves the best depth score, it is an image-only, training-free baseline and still exhibits noticeable edge artifacts in the stereo views (Fig.~\ref{fig:replica_supp}). 
In contrast, our method achieves competitive depth accuracy while producing cleaner stereo geometry.

\subsection{Failure Cases}
\label{sec:failure_cases}
\begin{figure*}[t]
\centering
\includegraphics[width=\textwidth]{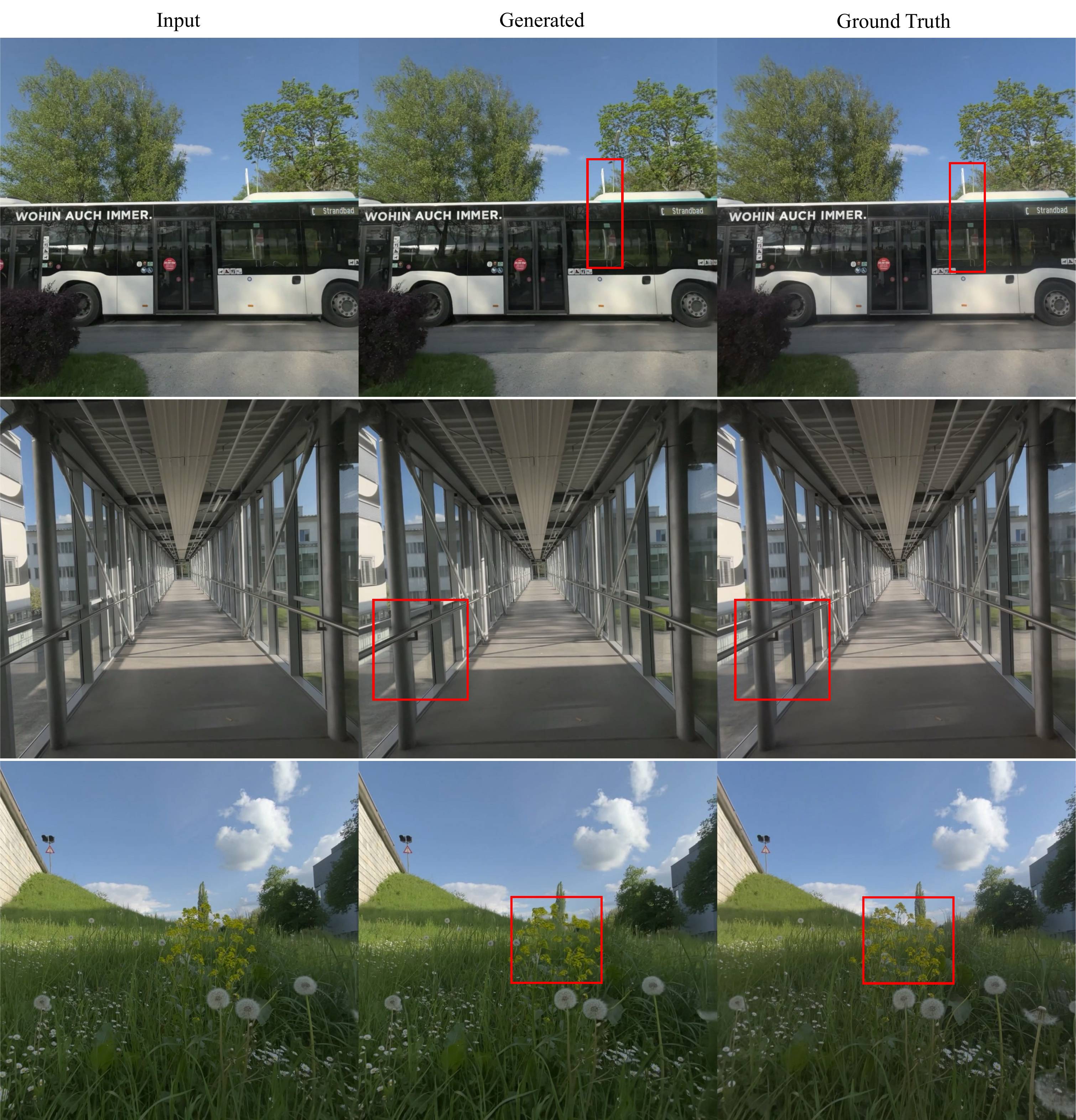}
\caption{Representative failure cases on transparent, reflective, and highly textured scenes from the SVD AVP test set.}
\label{fig:failure_cases}
\end{figure*}

Due to the lack of stronger priors for challenging real-world scenes, directly performing pixel-domain warping based on estimated depth still has inherent limitations. 
As shown in Figure~\ref{fig:failure_cases}, our method exhibits noticeable artifacts in several representative cases.
\begin{itemize}[leftmargin=1.2em]
    \item \textbf{Transparent objects.} Due to the geometric ambiguity introduced by transparency, the utility pole seen through the bus window becomes noticeably distorted.
    \item \textbf{Reflective surfaces.} The shadows reflected on the corridor glass are inconsistent with the true scene geometry, since the model cannot properly account for such view-dependent reflection effects.
    \item \textbf{High-frequency details.} The synthesized results appear insufficiently sharp, indicating limited fidelity in preserving fine-grained high-frequency textures.
\end{itemize}
These examples suggest that transparency, reflection, and dense high-frequency textures remain challenging for depth-based warping methods. 
We consider such cases as long-tail scenarios in view synthesis, which may be further alleviated in future work by improving geometric reasoning and incorporating stronger visual priors.

\clearpage


\clearpage
{
    \small
    \begin{thebibliography}{47}
\providecommand{\natexlab}[1]{#1}
\providecommand{\url}[1]{\texttt{#1}}
\expandafter\ifx\csname urlstyle\endcsname\relax
  \providecommand{\doi}[1]{doi: #1}\else
  \providecommand{\doi}{doi: \begingroup \urlstyle{rm}\Url}\fi

\bibitem[Blattmann et~al.(2023)Blattmann, Dockhorn, Kulal, Mendelevitch, Kilian, Lorenz, Levi, English, Voleti, Letts, et~al.]{blattmann2023stable}
Andreas Blattmann, Tim Dockhorn, Sumith Kulal, Daniel Mendelevitch, Maciej Kilian, Dominik Lorenz, Yam Levi, Zion English, Vikram Voleti, Adam Letts, et~al.
\newblock Stable video diffusion: Scaling latent video diffusion models to large datasets.
\newblock \emph{arXiv preprint arXiv:2311.15127}, 2023.

\bibitem[Dai et~al.(2024)Dai, Tan, Xu, Futschik, Du, Fanello, Qi, and Zhang]{dai2024svg}
Peng Dai, Feitong Tan, Qiangeng Xu, David Futschik, Ruofei Du, Sean Fanello, Xiaojuan Qi, and Yinda Zhang.
\newblock Svg: 3d stereoscopic video generation via denoising frame matrix.
\newblock \emph{arXiv preprint arXiv:2407.00367}, 2024.

\bibitem[{Depthify.ai}(2025)]{depthify}
{Depthify.ai}.
\newblock Depthify.ai.
\newblock \url{https://www.depthify.ai/}, 2025.
\newblock Accessed: 2025-07-31.

\bibitem[Gao et~al.(2024)Gao, Holynski, Henzler, Brussee, Martin-Brualla, Srinivasan, Barron, and Poole]{gao2024cat3d}
Ruiqi Gao, Aleksander Holynski, Philipp Henzler, Arthur Brussee, Ricardo Martin-Brualla, Pratul~P. Srinivasan, Jonathan~T. Barron, and Ben Poole.
\newblock Cat3d: Create anything in 3d with multi-view diffusion models.
\newblock \emph{Advances in Neural Information Processing Systems}, 2024.

\bibitem[Han et~al.(2022)Han, Wang, and Yang]{han2022single}
Yuxuan Han, Ruicheng Wang, and Jiaolong Yang.
\newblock Single-view view synthesis in the wild with learned adaptive multiplane images.
\newblock In \emph{ACM SIGGRAPH}, 2022.

\bibitem[Hong et~al.(2022)Hong, Peng, Xiao, Liu, and Zhang]{hong2022headnerf}
Yang Hong, Bo Peng, Haiyao Xiao, Ligang Liu, and Juyong Zhang.
\newblock Headnerf: A real-time nerf-based parametric head model.
\newblock In \emph{Proceedings of the IEEE/CVF Conference on Computer Vision and Pattern Recognition}, pages 20374--20384, 2022.

\bibitem[Hu et~al.(2025)Hu, Gao, Li, Zhao, Cun, Zhang, Quan, and Shan]{hu2025-DepthCrafter}
Wenbo Hu, Xiangjun Gao, Xiaoyu Li, Sijie Zhao, Xiaodong Cun, Yong Zhang, Long Quan, and Ying Shan.
\newblock Depthcrafter: Generating consistent long depth sequences for open-world videos.
\newblock In \emph{CVPR}, 2025.

\bibitem[Huang et~al.(2025)Huang, Singh, Dubost, Vasconcelos, Khattar, Shi, Theobalt, Oztireli, and Singh]{huang2025restereo}
Xingchang Huang, Ashish~Kumar Singh, Florian Dubost, Cristina~Nader Vasconcelos, Sakar Khattar, Liang Shi, Christian Theobalt, Cengiz Oztireli, and Gurprit Singh.
\newblock Restereo: Diffusion stereo video generation and restoration.
\newblock \emph{arXiv preprint arXiv:2506.06023}, 2025.

\bibitem[Izadimehr et~al.(2025)Izadimehr, Ghanbari, Chen, Zhou, Hao, Dasari, Timmerer, and Amirpour]{izadimehr2025svd}
M.H. Izadimehr, Milad Ghanbari, Guodong Chen, Wei Zhou, Xiaoshuai Hao, Mallesham Dasari, Christian Timmerer, and Hadi Amirpour.
\newblock Svd: Spatial video dataset.
\newblock In \emph{ACM International Conference on Multimedia (ACM MM)}, 2025.
\newblock Submitted.

\bibitem[Jiang et~al.(2025)Jiang, Han, Mao, Zhang, Pan, and Liu]{jiang2025vace}
Zeyinzi Jiang, Zhen Han, Chaojie Mao, Jingfeng Zhang, Yulin Pan, and Yu Liu.
\newblock Vace: All-in-one video creation and editing.
\newblock \emph{arXiv preprint arXiv:2503.07598}, 2025.

\bibitem[Jing et~al.(2024)Jing, Mao, and Mikolajczyk]{jing2024match}
Junpeng Jing, Ye Mao, and Krystian Mikolajczyk.
\newblock Match-stereo-videos: Bidirectional alignment for consistent dynamic stereo matching.
\newblock In \emph{European Conference on Computer Vision}, pages 415--432. Springer, 2024.

\bibitem[Karaev et~al.(2023)Karaev, Rocco, Graham, Neverova, Vedaldi, and Rupprecht]{karaev2023dynamicstereo}
Nikita Karaev, Ignacio Rocco, Benjamin Graham, Natalia Neverova, Andrea Vedaldi, and Christian Rupprecht.
\newblock Dynamicstereo: Consistent dynamic depth from stereo videos.
\newblock \emph{CVPR}, 2023.

\bibitem[Kerbl et~al.(2023)Kerbl, Kopanas, Leimk{\"u}hler, and Drettakis]{kerbl20233d}
Bernhard Kerbl, Georgios Kopanas, Thomas Leimk{\"u}hler, and George Drettakis.
\newblock 3d gaussian splatting for real-time radiance field rendering.
\newblock \emph{ACM Trans. Graph.}, 42\penalty0 (4):\penalty0 139--1, 2023.

\bibitem[Konrad et~al.(2013)Konrad, Wang, Ishwar, Wu, and Mukherjee]{konrad2013learning}
Janusz Konrad, Meng Wang, Prakash Ishwar, Chen Wu, and Debargha Mukherjee.
\newblock Learning-based, automatic 2d-to-3d image and video conversion.
\newblock \emph{IEEE Transactions on Image Processing}, 22\penalty0 (9):\penalty0 3485--3496, 2013.

\bibitem[Lv et~al.(2025)Lv, Long, Huang, Li, Lv, Ren, and Zheng]{2024spatialdreamer}
Zhen Lv, Yangqi Long, Congzhentao Huang, Cao Li, Chengfei Lv, Hao Ren, and Dian Zheng.
\newblock Spatialdreamer: Self-supervised stereo video synthesis from monocular input.
\newblock In \emph{Proceedings of the Computer Vision and Pattern Recognition Conference}, pages 811--821, 2025.

\bibitem[McGuire and McGuire(2005)]{mcguire2005steep}
Morgan McGuire and Max McGuire.
\newblock Steep parallax mapping.
\newblock \emph{I3D 2005 Poster}, pages 23--24, 2005.

\bibitem[Mehl et~al.(2024)Mehl, Bruhn, Gross, and Schroers]{mehl2024stereo}
Lukas Mehl, Andr{\'e}s Bruhn, Markus Gross, and Christopher Schroers.
\newblock Stereo conversion with disparity-aware warping, compositing and inpainting.
\newblock In \emph{Proceedings of the IEEE/CVF Winter Conference on Applications of Computer Vision}, pages 4260--4269, 2024.

\bibitem[Mildenhall et~al.(2021)Mildenhall, Srinivasan, Tancik, Barron, Ramamoorthi, and Ng]{mildenhall2021nerf}
Ben Mildenhall, Pratul~P Srinivasan, Matthew Tancik, Jonathan~T Barron, Ravi Ramamoorthi, and Ren Ng.
\newblock Nerf: Representing scenes as neural radiance fields for view synthesis.
\newblock \emph{Communications of the ACM}, 65\penalty0 (1):\penalty0 99--106, 2021.

\bibitem[Nan et~al.(2024)Nan, Xie, Zhou, Fan, Yang, Chen, Li, Yang, and Tai]{nan2024openvid}
Kepan Nan, Rui Xie, Penghao Zhou, Tiehan Fan, Zhenheng Yang, Zhijie Chen, Xiang Li, Jian Yang, and Ying Tai.
\newblock Openvid-1m: A large-scale high-quality dataset for text-to-video generation.
\newblock \emph{arXiv preprint arXiv:2407.02371}, 2024.

\bibitem[Niklaus and Liu(2020)]{Niklaus_CVPR_2020}
Simon Niklaus and Feng Liu.
\newblock Softmax splatting for video frame interpolation.
\newblock In \emph{IEEE Conference on Computer Vision and Pattern Recognition}, 2020.

\bibitem[{Owl3D}(2025)]{owl3d}
{Owl3D}.
\newblock owl3d.
\newblock \url{https://www.owl3d.com/landing}, 2025.
\newblock Accessed: 2025-07-31.

\bibitem[Peebles and Xie(2023)]{peebles2023scalable}
William Peebles and Saining Xie.
\newblock Scalable diffusion models with transformers.
\newblock In \emph{Proceedings of the IEEE/CVF international conference on computer vision}, pages 4195--4205, 2023.

\bibitem[Pumarola et~al.(2021)Pumarola, Corona, Pons-Moll, and Moreno-Noguer]{pumarola2021d}
Albert Pumarola, Enric Corona, Gerard Pons-Moll, and Francesc Moreno-Noguer.
\newblock D-nerf: Neural radiance fields for dynamic scenes.
\newblock In \emph{Proceedings of the IEEE/CVF conference on computer vision and pattern recognition}, pages 10318--10327, 2021.

\bibitem[Rombach et~al.(2021)Rombach, Blattmann, Lorenz, Esser, and Ommer]{rombach2021highresolution}
Robin Rombach, Andreas Blattmann, Dominik Lorenz, Patrick Esser, and Björn Ommer.
\newblock High-resolution image synthesis with latent diffusion models, 2021.

\bibitem[Shi et~al.(2024)Shi, Li, and Wonka]{shi2024immersepro}
Jian Shi, Zhenyu Li, and Peter Wonka.
\newblock Immersepro: End-to-end stereo video synthesis via implicit disparity learning.
\newblock \emph{arXiv preprint arXiv:2410.00262}, 2024.

\bibitem[Shvetsova et~al.(2025)Shvetsova, Bhat, Truong, Kuehne, and Tombari]{shvetsova2025m2svid}
Nina Shvetsova, Goutam Bhat, Prune Truong, Hilde Kuehne, and Federico Tombari.
\newblock M2svid: End-to-end inpainting and refinement for monocular-to-stereo video conversion.
\newblock \emph{arXiv preprint arXiv:2505.16565}, 2025.

\bibitem[Tucker and Snavely(2020)]{tucker2020single}
Richard Tucker and Noah Snavely.
\newblock Single-view view synthesis with multiplane images.
\newblock In \emph{Proceedings of the IEEE/CVF Conference on Computer Vision and Pattern Recognition}, pages 551--560, 2020.

\bibitem[Wan et~al.(2025)Wan, Wang, Ai, Wen, Mao, Xie, Chen, Yu, Zhao, Yang, Zeng, Wang, Zhang, Zhou, Wang, Chen, Zhu, Zhao, Yan, Huang, Feng, Zhang, Li, Wu, Chu, Feng, Zhang, Sun, Fang, Wang, Gui, Weng, Shen, Lin, Wang, Wang, Zhou, Wang, Shen, Yu, Shi, Huang, Xu, Kou, Lv, Li, Liu, Wang, Zhang, Huang, Li, Wu, Liu, Pan, Zheng, Hong, Shi, Feng, Jiang, Han, Wu, and Liu]{wan2025}
Team Wan, Ang Wang, Baole Ai, Bin Wen, Chaojie Mao, Chen-Wei Xie, Di Chen, Feiwu Yu, Haiming Zhao, Jianxiao Yang, Jianyuan Zeng, Jiayu Wang, Jingfeng Zhang, Jingren Zhou, Jinkai Wang, Jixuan Chen, Kai Zhu, Kang Zhao, Keyu Yan, Lianghua Huang, Mengyang Feng, Ningyi Zhang, Pandeng Li, Pingyu Wu, Ruihang Chu, Ruili Feng, Shiwei Zhang, Siyang Sun, Tao Fang, Tianxing Wang, Tianyi Gui, Tingyu Weng, Tong Shen, Wei Lin, Wei Wang, Wei Wang, Wenmeng Zhou, Wente Wang, Wenting Shen, Wenyuan Yu, Xianzhong Shi, Xiaoming Huang, Xin Xu, Yan Kou, Yangyu Lv, Yifei Li, Yijing Liu, Yiming Wang, Yingya Zhang, Yitong Huang, Yong Li, You Wu, Yu Liu, Yulin Pan, Yun Zheng, Yuntao Hong, Yupeng Shi, Yutong Feng, Zeyinzi Jiang, Zhen Han, Zhi-Fan Wu, and Ziyu Liu.
\newblock Wan: Open and advanced large-scale video generative models.
\newblock \emph{arXiv preprint arXiv:2503.20314}, 2025.

\bibitem[Wang et~al.(2024)Wang, Frisvad, Jensen, and Bigdeli]{wang2024stereodiffusion}
Lezhong Wang, Jeppe~Revall Frisvad, Mark~Bo Jensen, and Siavash~Arjomand Bigdeli.
\newblock Stereodiffusion: Training-free stereo image generation using latent diffusion models.
\newblock In \emph{Proceedings of the IEEE/CVF Conference on Computer Vision and Pattern Recognition}, pages 7416--7425, 2024.

\bibitem[Wang et~al.(2025)Wang, Yang, Xu, Cheng, Lin, Deng, Zang, Chen, and Yang]{wang2025zerostereo}
Xianqi Wang, Hao Yang, Gangwei Xu, Junda Cheng, Min Lin, Yong Deng, Jinliang Zang, Yurui Chen, and Xin Yang.
\newblock Zerostereo: Zero-shot stereo matching from single images.
\newblock \emph{arXiv preprint arXiv:2501.08654}, 2025.

\bibitem[Wang et~al.(2004)Wang, Bovik, Sheikh, and Simoncelli]{wang2004image}
Zhou Wang, Alan~C Bovik, Hamid~R Sheikh, and Eero~P Simoncelli.
\newblock Image quality assessment: from error visibility to structural similarity.
\newblock \emph{IEEE transactions on image processing}, 13\penalty0 (4):\penalty0 600--612, 2004.

\bibitem[Wen et~al.(2025{\natexlab{a}})Wen, Trepte, Aribido, Kautz, Gallo, and Birchfield]{wen2025foundationstereo}
Bowen Wen, Matthew Trepte, Joseph Aribido, Jan Kautz, Orazio Gallo, and Stan Birchfield.
\newblock Foundationstereo: Zero-shot stereo matching.
\newblock In \emph{Proceedings of the Computer Vision and Pattern Recognition Conference}, pages 5249--5260, 2025{\natexlab{a}}.

\bibitem[Wen et~al.(2025{\natexlab{b}})Wen, Trepte, Aribido, Kautz, Gallo, and Birchfield]{wen2025stereo}
Bowen Wen, Matthew Trepte, Joseph Aribido, Jan Kautz, Orazio Gallo, and Stan Birchfield.
\newblock Foundationstereo: Zero-shot stereo matching.
\newblock \emph{CVPR}, 2025{\natexlab{b}}.

\bibitem[{Wikipedia contributors}(2024)]{wikipedia2024psnr}
{Wikipedia contributors}.
\newblock Peak signal-to-noise ratio --- {Wikipedia}{,} the free encyclopedia.
\newblock \url{https://en.wikipedia.org/w/index.php?title=Peak_signal-to-noise_ratio&oldid=1210897995}, 2024.
\newblock [Online; accessed 4-March-2024].

\bibitem[Wu et~al.(2024)Wu, Yi, Fang, Xie, Zhang, Wei, Liu, Tian, and Wang]{wu20244d}
Guanjun Wu, Taoran Yi, Jiemin Fang, Lingxi Xie, Xiaopeng Zhang, Wei Wei, Wenyu Liu, Qi Tian, and Xinggang Wang.
\newblock 4d gaussian splatting for real-time dynamic scene rendering.
\newblock In \emph{Proceedings of the IEEE/CVF conference on computer vision and pattern recognition}, pages 20310--20320, 2024.

\bibitem[Xie et~al.(2016)Xie, Girshick, and Farhadi]{xie2016deep3d}
Junyuan Xie, Ross Girshick, and Ali Farhadi.
\newblock Deep3d: Fully automatic 2d-to-3d video conversion with deep convolutional neural networks.
\newblock In \emph{European conference on computer vision}, pages 842--857. Springer, 2016.

\bibitem[Xie et~al.(2024)Xie, Yao, Voleti, Jiang, and Jampani]{xie2024sv4d}
Yiming Xie, Chun-Han Yao, Vikram Voleti, Huaizu Jiang, and Varun Jampani.
\newblock Sv4d: Dynamic 3d content generation with multi-frame and multi-view consistency.
\newblock \emph{arXiv preprint arXiv:2407.17470}, 2024.

\bibitem[Yang et~al.(2024)Yang, Teng, Zheng, Ding, Huang, Xu, Yang, Hong, Zhang, Feng, et~al.]{yang2024cogvideox}
Zhuoyi Yang, Jiayan Teng, Wendi Zheng, Ming Ding, Shiyu Huang, Jiazheng Xu, Yuanming Yang, Wenyi Hong, Xiaohan Zhang, Guanyu Feng, et~al.
\newblock Cogvideox: Text-to-video diffusion models with an expert transformer.
\newblock \emph{arXiv preprint arXiv:2408.06072}, 2024.

\bibitem[YU et~al.(2025)YU, Hu, Xing, and Shan]{yu2025trajectorycrafter}
Mark YU, Wenbo Hu, Jinbo Xing, and Ying Shan.
\newblock Trajectorycrafter: Redirecting camera trajectory for monocular videos via diffusion models.
\newblock \emph{arXiv preprint arXiv:2503.05638}, 2025.

\bibitem[Yu et~al.(2024)Yu, Xing, Yuan, Hu, Li, Huang, Gao, Wong, Shan, and Tian]{yu2024viewcrafter}
Wangbo Yu, Jinbo Xing, Li Yuan, Wenbo Hu, Xiaoyu Li, Zhipeng Huang, Xiangjun Gao, Tien-Tsin Wong, Ying Shan, and Yonghong Tian.
\newblock Viewcrafter: Taming video diffusion models for high-fidelity novel view synthesis.
\newblock \emph{arXiv preprint arXiv:2409.02048}, 2024.

\bibitem[Zhang et~al.(2024)Zhang, Jia, Liu, Zhang, Wei, and Tian]{zhang2024spatialme}
Jiale Zhang, Qianxi Jia, Yang Liu, Wei Zhang, Wei Wei, and Xin Tian.
\newblock Spatialme: Stereo video conversion using depth-warping and blend-inpainting.
\newblock \emph{arXiv preprint arXiv:2412.11512}, 2024.

\bibitem[Zhang et~al.(2011)Zhang, Vazquez, and Knorr]{zhang20113d}
Liang Zhang, Carlos Vazquez, and Sebastian Knorr.
\newblock 3d-tv content creation: automatic 2d-to-3d video conversion.
\newblock \emph{IEEE Transactions on Broadcasting}, 57\penalty0 (2):\penalty0 372--383, 2011.

\bibitem[Zhang et~al.(2023)Zhang, Wang, Li, Huang, Sato, and Lu]{zhang2023structural}
Mingfang Zhang, Jinglu Wang, Xiao Li, Yifei Huang, Yoichi Sato, and Yan Lu.
\newblock Structural multiplane image: Bridging neural view synthesis and 3d reconstruction.
\newblock In \emph{Proceedings of the IEEE/CVF Conference on Computer Vision and Pattern Recognition}, pages 16707--16716, 2023.

\bibitem[Zhang et~al.(2018)Zhang, Isola, Efros, Shechtman, and Wang]{zhang2018unreasonable}
Richard Zhang, Phillip Isola, Alexei~A Efros, Eli Shechtman, and Oliver Wang.
\newblock The unreasonable effectiveness of deep features as a perceptual metric.
\newblock In \emph{Proceedings of the IEEE conference on computer vision and pattern recognition}, pages 586--595, 2018.

\bibitem[Zhao et~al.(2024{\natexlab{a}})Zhao, Hu, Cun, Zhang, Li, Kong, Gao, Niu, and Shan]{zhao2024stereocrafter}
Sijie Zhao, Wenbo Hu, Xiaodong Cun, Yong Zhang, Xiaoyu Li, Zhe Kong, Xiangjun Gao, Muyao Niu, and Ying Shan.
\newblock Stereocrafter: Diffusion-based generation of long and high-fidelity stereoscopic 3d from monocular videos.
\newblock \emph{arXiv preprint arXiv:2409.07447}, 2024{\natexlab{a}}.

\bibitem[Zhao et~al.(2024{\natexlab{b}})Zhao, Zhang, Cun, Yang, Niu, Li, Hu, and Shan]{zhao2024cvvae}
Sijie Zhao, Yong Zhang, Xiaodong Cun, Shaoshu Yang, Muyao Niu, Xiaoyu Li, Wenbo Hu, and Ying Shan.
\newblock Cv-vae: A compatible video vae for latent generative video models.
\newblock \emph{Advances in Neural Information Processing Systems}, 37:\penalty0 12847--12871, 2024{\natexlab{b}}.

\bibitem[Zhou et~al.(2023)Zhou, Li, Chan, and Loy]{zhou2023propainter}
Shangchen Zhou, Chongyi Li, Kelvin~CK Chan, and Chen~Change Loy.
\newblock Propainter: Improving propagation and transformer for video inpainting.
\newblock In \emph{Proceedings of the IEEE/CVF international conference on computer vision}, pages 10477--10486, 2023.

\end{thebibliography}

}
\end{document}